\documentclass[letterpaper]{article} 

\usepackage{aaai23}  

\usepackage{times}  
\usepackage{helvet}  
\usepackage{courier}  
\usepackage[hyphens]{url}  
\usepackage{graphicx} 
\usepackage{natbib}  
\usepackage{caption} 
\usepackage[utf8]{inputenc} 
\usepackage[T1]{fontenc}    
\usepackage{hyperref}       
\usepackage{url}            
\usepackage{booktabs}       
\usepackage{amsfonts}       
\usepackage{nicefrac}       
\usepackage{microtype}      
\usepackage{xcolor}         

\usepackage{algorithm}
\usepackage{comment}

\usepackage{caption}

\usepackage{microtype}
\usepackage{graphicx}

\usepackage{caption}
\usepackage{subcaption}
\usepackage{breqn}

\usepackage{booktabs} 

\usepackage{amsmath}
\usepackage{amssymb}
\usepackage{bbm}
\usepackage{amsthm}
\usepackage{threeparttable}
\usepackage{array}
\usepackage{graphicx}
\usepackage{makecell}
\usepackage[capitalise]{cleveref}

\usepackage{xcolor}
\usepackage{csquotes}
\usepackage{wrapfig}
\usepackage{varwidth}
\usepackage{capt-of}
\usepackage{multicol}

\usepackage{caption}
\DeclareCaptionLabelFormat{andfigure}{#1 #2 \& \figurename{} \thefigure{}}

\definecolor{ourspecialtextcolor}{rgb}{0.528, 0.471, 0.701} 

\usepackage{algorithm}
\usepackage[noend]{algpseudocode}
\DeclareCaptionSubType*{algorithm}

\algrenewcommand{\algorithmiccomment}[1]{\bgroup\hfill//~#1\egroup}
\algrenewcommand{\Return}{\State\textbf{return}\ }
\algnewcommand{\Save}{\State\textbf{save}\ }
\algnewcommand{\Load}{\State\textbf{load}\ }

\algnewcommand{\LeftComment}[1]{\Statex \(\triangleright\) #1}

\DeclareMathOperator{\fMAP}{\mathtt{MAP}}

\DeclareMathOperator{\GumbelDist}{Gumbel}

\usepackage{bm}
\DeclareMathOperator*{\argmax}{arg\,max}

\usepackage{mathtools}

\usepackage{multirow}

\usepackage{xspace}
\usepackage{adjustbox}

\usepackage{placeins}

\newcommand{\bz}{\bm{z}}

\newcommand{\bv}{\bm{v}}

\newcommand{\bepsilon}{\bm{\epsilon}}

\newcommand{\bg}{\bm{g}}

\newcommand{\bmu}{\bm{\mu}}
\newcommand{\btheta}{\bm{\theta}}

\newcommand{\grad}[2]{\nabla_{#1}#2}

\newcommand{\distparamspace}{\Theta}
\newcommand{\latentprobdist}{p(\bz; \btheta)}

\newcommand{\bnoisedist}{\rho(\bepsilon)}

\usepackage{paralist}
\usepackage{wasysym}
\usepackage{marvosym}

\usepackage{xspace}
\newcommand*{\eg}{e.g.\@\xspace}
\newcommand*{\ie}{i.e.\@\xspace}

\newcommand{\imle}{\textsc{IMLE}\@\xspace}
\newcommand{\aimle}{\textsc{AIMLE}\@\xspace}

\newcommand{\norm}[1]{\left\lVert#1\right\rVert}

\usepackage{etoolbox}

\newbool{proofsIMLE}

\newbool{printOld}

\newbool{printChecklist}

\newbool{printApdx}
\booltrue{printApdx}
\booltrue{printApdx}

\newbool{extraspace}

\begin{document}

\author{Pasquale Minervini$^{\text{\Leo \Aries}}$ \qquad Luca Franceschi$^{\text{\Aries}}$ \qquad Mathias Niepert$^{\text{\Taurus}}$  \\
\normalfont $^{\text{\Leo}}$ School of Informatics, University of Edinburgh, Edinburgh, United Kingdom \\
$^{\text{\Aries}}$ UCL Centre for Artificial Intelligence, London, United Kingdom \\
$^{\text{\Taurus}}$ University of Stuttgart, Stuttgart, Germany
}

\title{Adaptive Perturbation-Based Gradient Estimation \\ for Discrete Latent Variable Models}

\maketitle

\begin{abstract}
The integration of discrete algorithmic components in deep learning architectures has numerous applications.
Recently, Implicit Maximum Likelihood Estimation~\citep[IMLE,][]{niepert21imle}, a class of gradient estimators for discrete exponential family distributions, was proposed by combining implicit differentiation through perturbation with the path-wise gradient estimator.
However, due to the finite difference approximation of the gradients, it is especially sensitive to the choice of the finite difference step size, which needs to be specified by the user. 
In this work, we present Adaptive IMLE (\aimle), the first adaptive gradient estimator for complex discrete distributions: it adaptively identifies the target distribution for \imle by trading off the density of gradient information with the degree of bias in the gradient estimates.
We empirically evaluate our estimator on synthetic examples, as well as on Learning to Explain, Discrete Variational Auto-Encoders, and Neural Relational Inference tasks.
In our experiments, we show that our adaptive gradient estimator can produce faithful estimates while requiring orders of magnitude fewer samples than other gradient estimators.
All source code and datasets are available at \url{https://github.com/EdinburghNLP/torch-adaptive-imle}.
\end{abstract}

\section{Introduction}

\begin{figure}[t!]
\centering
\includegraphics[width=\columnwidth]{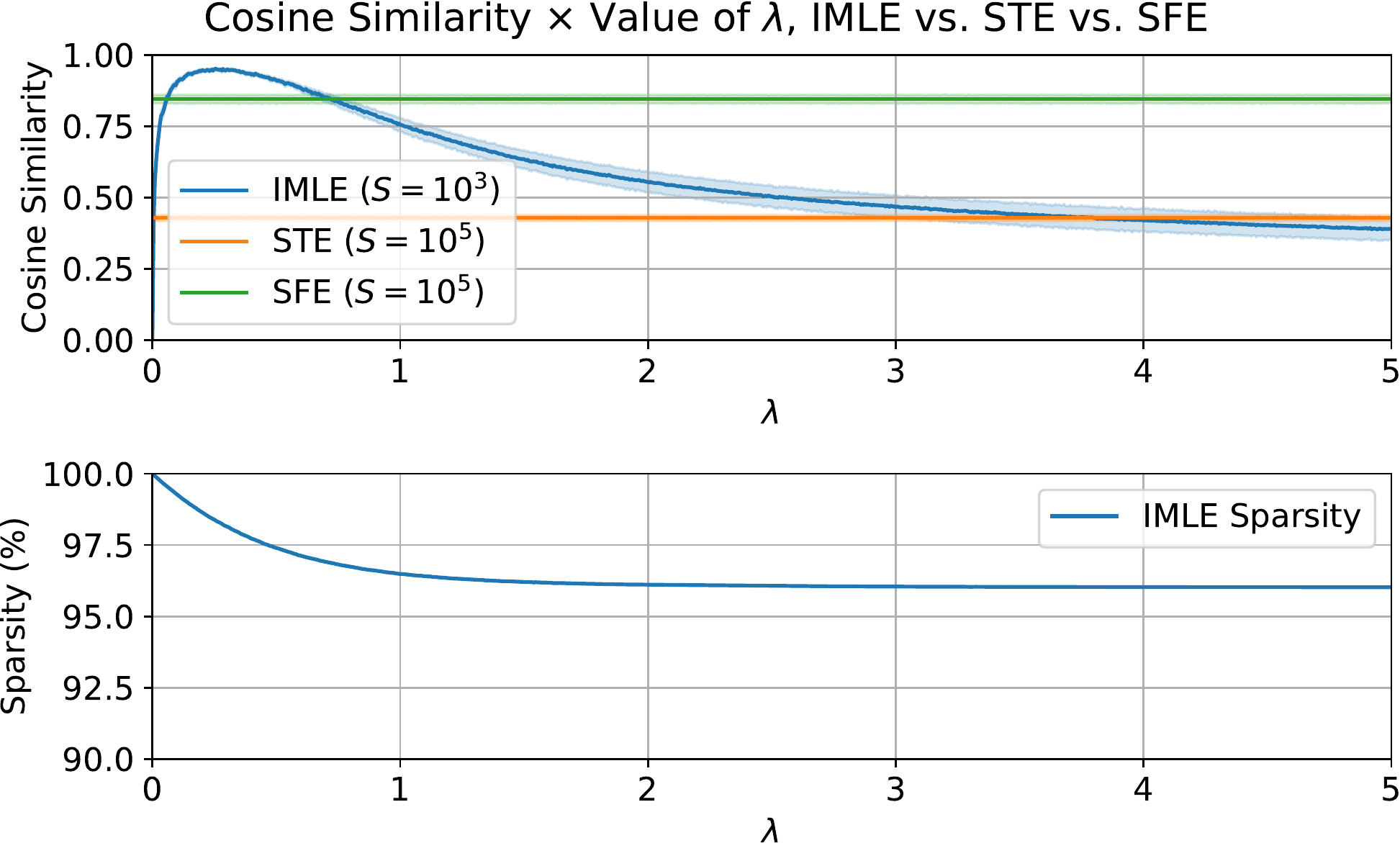}
\caption{(Top) cosine similarity between the true and estimated gradients $\nabla_{\theta} \mathbb{E}_{\bz \sim p(\bz; \btheta)} [ \norm{\bz - \mathbf{b}}^2 ]$, with $\mathbf{b} \sim \mathcal{N}(0,\mathbf{I})$ and $\bz = \{0, 1\}^{50}$ such that $\sum_{i} \bz_{i} = 1$, where estimates are computed using \imle~\citep{niepert21imle} with $S = 10^{3}$ samples, the Straight-Through Estimator~\citep[STE,][]{bengio2013estimating} with $S = 10^{5}$ samples, and the Score Function Estimator~\citep[SFE,][]{DBLP:journals/ml/Williams92} with $S = 10^{5}$ samples, and (bottom) sparsity (\% of zero elements) of the estimate \imle gradient -- results were averaged across 128 seeds.
For $\lambda \to 0$, the \imle gradient estimate is $\mathbf{0}$, while increasing $\lambda$ leads to increasingly more biased gradient estimates.
}
\label{fig:intro}
\end{figure}

\begin{figure}
  \centering
  \includegraphics[width=\columnwidth]{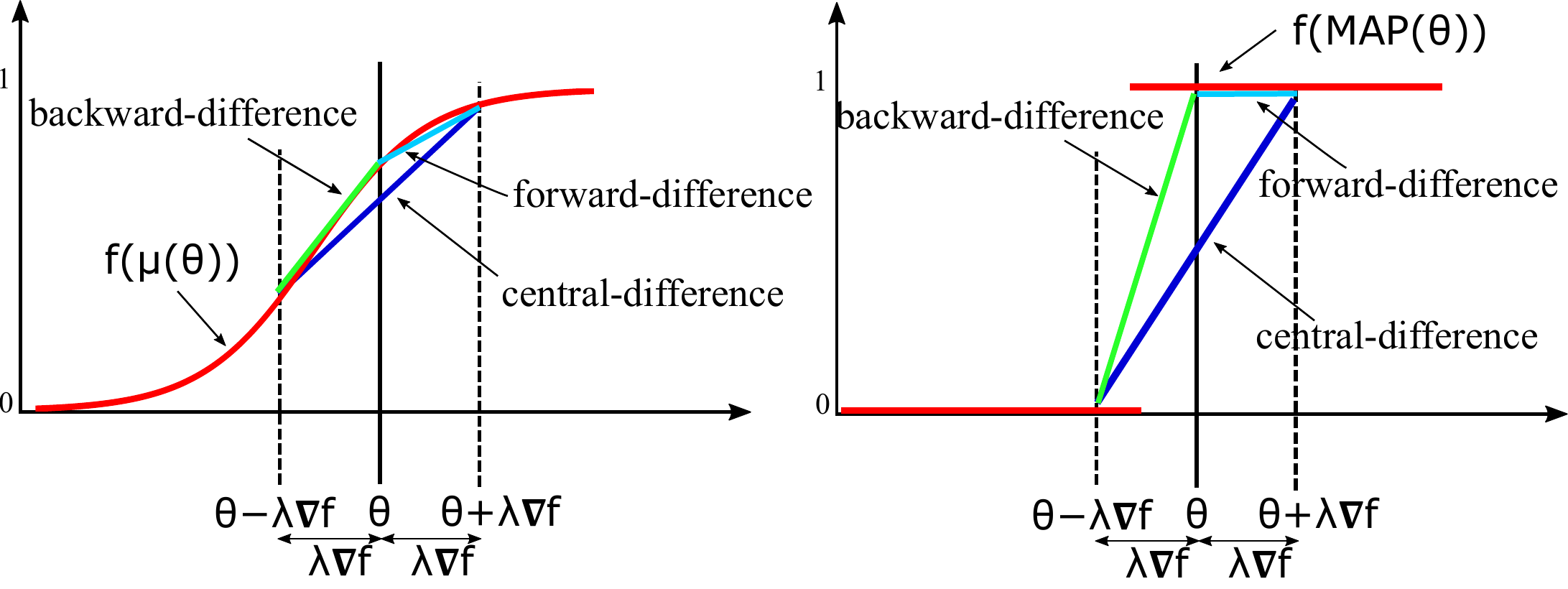}
  \caption{Finite difference approximation of a downstream function $f$ on continuous marginals (left) and discrete samples (right). The step size $\lambda$ trades off bias and sparsity of the gradient approximations for discrete samples and we propose to make the step size $\lambda$ adaptive.} \label{fig:fdgd}
\end{figure}

There is a growing interest in end-to-end learnable models incorporating discrete algorithms that allow, \eg to sample from discrete latent distributions~\citep{jang2016categorical,paulus2020gradient} or solve combinatorial optimisation problems~\citep{poganvcic2019differentiation,Mandi_Guns:2020,niepert21imle}.
These discrete components are not continuously differentiable, and an important problem is to efficiently estimate the gradients of their inputs to perform backpropagation. 
Reinforcement learning, discrete Energy-Based Models~\citep[EBMs,][]{lecun2006tutorial}, learning to explain~\cite{chen2018learning}, discrete Variational Auto-Encoders~\citep[VAEs,][]{DBLP:journals/corr/KingmaW13}, and discrete world models~\citep{hafner2020mastering} are additional examples of neural network-based architectures that require the ability to back-propagate through expectations of discrete probability distributions.
The main challenge these approaches have in common is the problem of (approximately) computing gradients of an expectation of a continuously differentiable function $f$:
\begin{equation}
 \nabla_{\btheta} \mathbb{E}_{\bz \sim p(\bz; \btheta)}[f(\bz)],
\end{equation}
where the expectation is taken over a complex discrete probability distribution with intractable marginals and normalisation constant.
In principle, one could use the Score Function Estimator \citep[SFE,][]{DBLP:journals/ml/Williams92}.
Unfortunately, it suffers from high variance typically exacerbated by the distribution $p(\bz; \btheta)$ being intractable.
Implicit Maximum Likelihood Estimation~\citep[\imle,][]{niepert21imle}, a recently proposed general-purpose gradient estimation technique, has shown lower variance and outperformed other existing methods, including the score function estimator and problem-specific continuous relaxations, in several settings~\cite{niepert21imle,betz:2021,qian2022ordered}.
For instance, for the synthetic problem in \cref{fig:intro}, the gradient estimate produced by SFE based on $10^5$ samples is worse than the estimate based on $10^3$ --- two orders of magnitude fewer --- samples using \imle due to the high variance of the SFE.
\imle combines Perturb-and-MAP sampling with a finite difference method for implicit differentiation originally developed for loss functions defined over marginals~\cite{domke:2010}.
In \imle, gradients are approximated as:
\begin{align}
 \hspace{-2mm} & \nabla_{\btheta} \mathbb{E}_{\bz\sim p(\bz; \btheta)}\left[f(\bz)\right]  \approx \notag \\
 & \label{eqn-imle-sample} \sum_{i=1}^{n} \frac{1}{\lambda} \biggl\{ \fMAP(\btheta + \bepsilon_i) - \fMAP\left(\btheta + \bepsilon_i - \lambda \grad{\bz_i}f(\bz_i) \right) \biggr\},
\end{align}
where $\fMAP(\btheta)$ is a maximum-probability state of the distribution $p(\bz; \btheta)$, $\bepsilon_i \sim \bnoisedist$ is a perturbation drawn from a noise distribution, and $\bz_i = \fMAP(\btheta + \bepsilon_i)$.
Computing MAP the states $\fMAP(\btheta)$ instead of sampling from $p(\bz;\theta)$ is especially interesting since, in many cases, it has lower computational complexity than sampling the corresponding distribution~\citep{niepert21imle}.
Crucially, the parameter $\lambda$ determines the step size of the finite difference approximation. When the input to $f$ is $p(\bz; \btheta)$'s continuously differentiable marginals, we have that smaller values of $\lambda$ lead to less biased estimates.
Hence, in this setting, $\lambda$ is typically set to a value that depends on the machine precision to void numerical instabilities~\cite{domke:2010}.
In the setting, we consider, however, that the input to $f$ is discrete and discontinuous.
Setting $\lambda$ to a very small value, in this case, results in zero gradients. This is illustrated in \cref{fig:fdgd} (right) for the forward difference method.
Hence, the crucial insight is that $\lambda$ trades off the bias and sparsity of the gradient approximation.
In \cref{fig:intro} (top) and (bottom), we plot, respectively, the bias and the sparsity of the gradient estimates for different values of $\lambda$ on a toy optimisation problem.
As we can see, larger values of $\lambda$ result in a higher bias, and low values of $\lambda$ result in gradient estimates almost always being zero.
With this paper, we propose to make the parameter $\lambda$ adaptive.
We also provide empirical results showing that making $\lambda$ adaptive reduces the bias and improves the results on several benchmark problems. 
\section{Problem Definition}
We consider the problem of computing the gradients of an expectation over a discrete probability distribution of a continuously differentiable function $f$, that is, 
\begin{equation}
\label{eq:hybridm}
 \nabla_{\btheta} \mathbb{E}_{\bz \sim p(\bz; \btheta)}[f(\bz)]
\end{equation}
where $\latentprobdist$ is a discrete probability distribution over binary vectors $\bz$ and with parameters $\btheta$.
Specifically, we are concerned with settings where $p(\bz; \btheta)$ is a discrete probability distribution with an intractable normalisation constant. 
Moreover, we assume that the function $f$ is a parameterized non-trivial continuously differentiable function which makes existing approaches such as direct loss minimisation and perturbed optimizers~\citep{berthet2020learning} not directly applicable.
More formally, let $\btheta \in \distparamspace \subseteq \mathbb{R}^m$ be a real-valued parameter vector.
The probability mass function (PMF) of a discrete constrained exponential family r.v. is:
\begin{equation} \label{def-constrained-exp-family}
p(\bz; \btheta) = \left\lbrace
\begin{array}{ll}
     \exp\left(\langle\bz,\btheta\rangle - A(\btheta) \right) & \text{if } \bz \in \mathcal{C}, \\
     0 & \text{otherwise.}
\end{array}
\right.
\end{equation}
Here, $\langle\cdot, \cdot\rangle$ is the inner product.
$A(\btheta)$ is the log-partition function, defined as $A(\btheta) = \log\left(\sum_{\bz \in \mathcal{C}} \exp \left(\langle\bz,\btheta\rangle  \right)\right)$, and $\mathcal{C}$ is an integral polytope of feasible configurations $\bz$.
We call $\langle \bz, \btheta\rangle$ the \emph{weight} of the state $\bz$.
The \emph{marginals} (expected value, mean) of 
the r.v.s $\mathbf{Z}$ are defined as $\bmu(\btheta) \coloneqq \mathop{\mathbb{E}}_{\bz \sim \latentprobdist}[\bz]$. 
Finally, the most probable states also referred to as the \emph{Maximum A-Posteriori} (MAP) states, are defined as $\fMAP(\btheta) \coloneqq \argmax_{\bz \in \mathcal{C}} \;\langle\bz,\btheta\rangle$.

\section{\textsc{AIMLE}: Adaptive Implicit Maximum-Likelihood Learning}

We base our 
estimator on a finite difference method for implicit differentiation~\cite{domke:2010,niepert21imle}, which is generally applicable to any discrete distribution as defined in \cref{def-constrained-exp-family}.
For the initial derivation, we assume that we can compute exact samples from the distribution $p(\bz; \btheta)$ using Perturb-and-MAP~\citep{Papandreou:2011} with noise distribution $\bnoisedist$. We write $\btheta + \bepsilon$ for a perturbation of the parameters $\btheta$ by a sample $\bepsilon$ from a noise distribution $\rho(\bepsilon)$.
Since, in general, this is not possible for complex distributions, using approximate Perturb-and-MAP samples introduces a bias in the gradient estimates.
For now, however, we assume that these perturbations are exact, \ie, that $\fMAP(\btheta +  \bepsilon) \sim p(\bz; \btheta)$.
Under these assumptions, and by invoking the law of the unconscious statistician~\cite{mohamed2019monte}, we obtain:
\begin{equation*}
\nabla_{\btheta} \mathbb{E}_{\bz\sim p(\bz; \btheta)}\left[f(\bz)\right] = \nabla_{\btheta}\mathbb{E}_{\bepsilon \sim \bnoisedist} [f(\fMAP(\btheta + \bepsilon))].
\end{equation*}
We now approximate $\fMAP(\btheta + \epsilon)$ by $\mu\left(\frac{\btheta + \bepsilon}{\tau}\right)$ for a small $\tau > 0$.
That is, we replace a MAP state with a corresponding vector representing the marginal probabilities where we can make the probabilities increasingly spikier by lowering a temperature parameter $\tau$. 
The approximation error between these two terms can be made arbitrarily small since:
\begin{equation*}
\mathbb{E}_{\bepsilon \sim \bnoisedist}\left[ f(\fMAP(\btheta + \epsilon)) \right] = \mathbb{E}_{\bepsilon \sim \bnoisedist}\left[ f\left(\lim_{\tau \rightarrow 0}\mu\left(\frac{\btheta + \bepsilon}{\tau}\right)\right) \right].
\end{equation*}
The above equality holds almost everywhere if the noise distribution is such that the probability of two or more components of $\btheta + \bepsilon$ being equal is zero.
This is the case in the standard setting where  $\rho(\bepsilon)\sim \mathrm{Gumbel}(0, 1)$~\citep{Papandreou:2011}.
Therefore, we can write that, for some $\tau > 0$,
\begin{equation*}
\nabla_{\btheta} \mathbb{E}_{\bepsilon \sim \bnoisedist}\left[ \fMAP(\btheta + \epsilon) \right]  \approx \grad{\btheta} \mathbb{E}_{\bepsilon \sim \bnoisedist} \left[f\left( \mu\left(\frac{\btheta + \bepsilon}{\tau}\right)\right)\right].
\end{equation*}
Writing the expectation as an integral, we have:
\begin{equation}
\begin{aligned}
& \grad{\btheta} \mathbb{E}_{\bepsilon \sim \bnoisedist} \left[f\left( \mu\left(\frac{\btheta + \bepsilon}{\tau}\right)\right)\right] \\
& \qquad =  \grad{\btheta} \int_{\mathbb{R}} p(\epsilon) f\left( \mu\left(\frac{\btheta + \bepsilon}{\tau}\right)\right) d\bepsilon.
\end{aligned}
\end{equation}
Now we can exchange differentiation and integration since, for finite $\tau>0$, $f$ and $\mu$ are continuously differentiable:
\begin{align}
& \grad{\btheta} \mathbb{E}_{\bepsilon \sim \bnoisedist} \left[f\left( \mu\left(\frac{\btheta + \bepsilon}{\tau}\right)\right)\right] \\
& \qquad =  \int_{\mathbb{R}} p(\epsilon) \grad{\btheta} f\left( \bmu\right) d\epsilon \ \  \text{with} \ \ \ \bmu := \mu\left(\frac{\btheta + \bepsilon}{\tau}\right) \notag \\
& \qquad = \mathbb{E}_{\bepsilon \sim \bnoisedist} \left[ \grad{\btheta} f\left( \bmu\right)\right] \notag \\
& \qquad = \mathbb{E}_{\bepsilon \sim \bnoisedist} \left[ \lim_{\lambda \rightarrow 0} \frac{1}{\lambda} \left\{ \bmu - \mu\left(\frac{\btheta + \bepsilon}{\tau} - \lambda \grad{\bmu}f(\bmu)\right)\right\} \right]. \notag
\end{align}
The last equality uses implicit differentiation by perturbation~\cite{domke:2010}, which is a finite difference method for computing the gradients of a function defined on marginals.
Finally, we again approximate the expression $\bmu = \mu\left(\frac{\btheta + \bepsilon}{\tau}\right)$ with $\bz := \fMAP(\btheta + \epsilon)$ and obtain:
\begin{align}
& \nabla_{\btheta} \mathbb{E}_{\bz\sim p(\bz; \btheta)}\left[f(\bz)\right]  \approx \label{imle-main-equation} \\
& \qquad \mathbb{E}_{\bepsilon \sim \bnoisedist} \biggl[ \lim_{\lambda \rightarrow 0} \frac{1}{\lambda} \biggl\{ \bz - \fMAP\left(\btheta + \bepsilon - \lambda \grad{\bz}f(\bz) \right) \biggr\} \biggr]. \notag
\end{align}
Again, the approximation error of the above expression is arbitrarily small (but not zero) because the derivation shown here is valid for any $\tau > 0$. A finite sample approximation of \cref{imle-main-equation} results in the \imle gradient estimator of \cref{eqn-imle-sample} given in the introduction.
While we could, in principle, use the marginals as input to the function $f$ as in relaxed gradient estimators~\cite{maddison2016concrete,jang2016categorical,paulus2020gradient}, computing marginals for the complex distributions we consider here is not tractable in general, and we have to use approximate Perturb-and-MAP samples.
\algrenewcommand\algorithmicindent{1em}
\algrenewcommand{\algorithmiccomment}[1]{\bgroup\hskip1em\textcolor{ourspecialtextcolor}{//~\textsl{#1}}\egroup}

\begin{algorithm*}[t]
\begin{multicols}{2}
\begin{algorithmic}
       \Function{Init}{}
       \State $\alpha \leftarrow 0$ \quad \ \Comment{Initial value of $\alpha$}
       \State $\overline{g} \leftarrow 1$ \quad \ \Comment{Initial gradient norm estimate}
       \State $\eta \leftarrow 10^{-3}$ \Comment{Update step for $\alpha$}
       \EndFunction
       
       \item[]
       
       \Function{ForwardPass}{$\btheta$}
       \\ \Comment{Sample from the noise distribution $\rho(\bepsilon)$}
       \State $\bepsilon_{1}, \ldots, \bepsilon_{N} \sim \rho(\bepsilon)$
       \\ \Comment{MAP states of perturbed $\btheta$}
       \State $\bz_{i} \leftarrow \fMAP(\btheta + \bepsilon_{i})$, for $i = 1, \ldots, N$
       \Save $\btheta$, $\bepsilon_{1}, \ldots, \bepsilon_N$, and $\bz_{1}, \ldots, \bz_{N}$ 
       \Return $\bz_{1}, \ldots, \bz_{N}$
       \EndFunction
       
       \item[]
\end{algorithmic}
\begin{algorithmic}
       \Function{BackwardPass}{$\grad{\bz_1} f(\bz_{1}), \ldots, \grad{\bz_N}f(\bz_{N})$}
       \Load $\btheta$, $\bepsilon_{1}, \ldots, \bepsilon_N$, and $\bz_{1}, \ldots, \bz_N$ 
       
       \State $\lambda = \alpha \frac{1}{N} \sum_{i=1}^{N} \frac{ \norm{\btheta}_{2} }{ \norm{\grad{\bz_i}f(\bz_{i})}_{2} }$ (see \cref{eq:lambda})
       
       
       
       \For{$i = 1, \ldots, N$}
       
           \State $\btheta_{R_{i}}' \leftarrow \btheta - \lambda \grad{\bz_i}f(\bz_{i})$
           \State $\btheta_{L_{i}}' \leftarrow \btheta + \lambda \grad{\bz_i}f(\bz_{i})$
           
           
           \State $\bm{g}_i \leftarrow \frac{1}{2 \lambda} \left[ \fMAP(\btheta_{L_{i}}' + \bepsilon_{i}) - \fMAP(\btheta_{R_{i}}' + \bepsilon_{i}) \right]$
       \EndFor
       
       \\ \Comment{Moving average of the gradient norm}
       \State $\overline{g} \leftarrow 0.9 \ \overline{g} + 0.1 \ \frac{1}{N} \sum_{i = 1}^{N} \norm{\bg_{i}}_{0}$
       
       \\ \Comment{Update $\alpha$ to make $\overline{g}$ closer to $c$}
       \State $\alpha \leftarrow \left[ \alpha + \left( \eta \text{ if } \ \overline{g} \leq c \ \text{else} - \eta \right) \right]_{+}$
       
       \Return $\frac{1}{N}\sum_{i = 1}^{N} \bg_{i}$
       \EndFunction
\end{algorithmic}
\end{multicols}
\caption{Central Difference Perturbation-based 
 Adaptive Implicit Maximum Likelihood Estimation (\aimle).}
\label{algo:main}
\end{algorithm*}

\subsection{An Adaptive Optimiser for Finite-Difference based Implicit Differentiation} \label{section-imle-framework}
An important observation that motivates the proposed adaptive version of \imle is that we need to choose a hyperparameter $\lambda \in \mathbb{R}_{+}$ for \cref{eqn-imle-sample}.
Choosing a very small $\lambda$ leads to most gradients being zero.
Consequently, the gradients being back-propagated to the upstream model $f_{\bv}$ are zeros, which prevents the upstream model from being trained.
If we choose $\lambda$ too large, we obtain less sparse gradients, but the gradients are also more biased.
Hence, we propose an optimiser that adapts $\lambda$ during training to trade off non-zero but biased and sparse but unbiased gradients.
Similar to adaptive first-order optimizers in deep learning, we replace a single hyperparameter with a set of new ones but show that we obtain consistently better results when using default hyperparameters for the adaptive method.
\paragraph{Normalisation of the perturbation strength.}
Our first observation is that the magnitude of the perturbation $\lambda$ in the direction of the negative downstream gradient in \cref{imle-main-equation}  highly depends on $\btheta$, the gradients of the downstream function $f$.
To mitigate the variations in the downstream gradients norm relative to the parameters $\btheta$, we propose to set a perturbation magnitude (the norm of the difference between $\btheta$ and the perturbed $\btheta$) to be a fraction of the norm of the parameter vector $\btheta$.
In particular, let $\alpha \geq 0$ be such a fraction, then we seek $\lambda$ such that:
\begin{equation} \label{eq:lambda}
\begin{aligned}
    \norm{\btheta - \btheta'}_{2} = \alpha \norm{\btheta}_{2} & \Leftrightarrow \lambda \norm{\nabla_z f(z)}_{2} = \alpha \norm{\btheta}_{2} \\
    & \Leftrightarrow \lambda = \alpha
    \frac{\norm{\btheta}_{2}}{\norm{\nabla_z f(z)}_{2}}.
\end{aligned}
\end{equation}
This way, we ensure that a global value for $\lambda$  roughly translates to the same input-specific magnitude of the perturbation in the direction of the negative gradient. 

\paragraph{Trading off Bias and Sparsity of the Gradient Estimates.}
For computing $\lambda$ as in \cref{eq:lambda}, we track the sparsity of the gradient estimator with an exponential moving average of the gradient norm.
Since the gradients -- \ie the difference between the two MAP states -- in \cref{eqn-imle-sample} for each $i$ are always in $\{-1, 0, 1\}$, we take the $L_0$-norm which is here equivalent to the number of non-zero gradients.
Consider a batch of $N$ inputs during training. 
Let $\hat{\nabla}_{\btheta}(j) := \fMAP(\btheta_j + \bepsilon_j) - \fMAP(\btheta_j + \bepsilon_j - \lambda \grad{\bz_j}f(\bz_j))$ with $\bz_j = \fMAP(\btheta_j + \bepsilon_j)$ be a single-sample gradient estimate from \cref{eqn-imle-sample} for an input data point $j \in \{1, ..., N\}$ with input parameters $\btheta_j$ and without the scaling factor $1/\lambda$. 
We compute, in every training iteration $t > 0$ and using a discount factor $0 < \gamma \leq 1$, the exponential moving average of the number of non-zero gradients per training example:
\begin{equation}
    \begin{aligned}
        \overline{g}_{t+1} = \gamma \overline{g}_{t} + (1 - \gamma) \frac{1}{N}\sum_{j=1}^{N} \norm{\hat{\nabla}_{\btheta}(j)}_{0}.
        \end{aligned}
\end{equation}
Similarly to adaptive optimisation algorithms for neural networks, we introduce an update rule for $\lambda$.
Let $c$ be the desired learning rate, expressed as the \emph{number of non-zero gradients per example}.
This is the target learning rate, that is, the desired number of gradients we obtain on average per example.
A typical value is $c = 1.0$, meaning that we aim to adapt the value for $\lambda$ to obtain, on average, at least one non-zero gradient per example. 
We now use the following update rule for some fixed $\eta > 0$:
\begin{equation} 
\alpha_{t + 1} = \left\lbrace
\begin{array}{ll}
     \alpha_{t} + \eta & \text{if } \overline{g}_{t+1} \leq c, \\
     \alpha_{t} - \eta & \text{otherwise.} 
\end{array}
\right.
\end{equation}
Hence, by increasing or decreasing $\alpha$ by a constant factor and based on the current exponential moving average of the gradient sparsity, we adapt $\lambda$ through \cref{eq:lambda}, which relates $\lambda$ and $\alpha$.
\cref{algo:main} lists the gradient estimator as a layer in a neural network with a forward and backward pass. 

\paragraph{Forward and centred finite difference approximation.}
Gradient estimation in \imle, as outlined in \cref{eqn-imle-sample}, is analogous to gradient estimation with forward (one-sided) finite difference approximations, where $\left[ g(x + h) - g(x) \right] / h \approx g'(x)$.
A better approximation can be obtained by the \emph{centred} (two-sided) \emph{difference formula} $\left[ g(x + h) - g(x - h) \right] / 2h$, which is a second-order approximation to the first derivative~\citep{Olver2013-sj}.
Following this intuition, we replace $\frac{1}{\lambda} \left[ \fMAP\left(\btheta + \bepsilon \right) - \fMAP\left(\btheta + \bepsilon - \lambda \grad{\bz}f(\bz) \right) \right]$ in \cref{eqn-imle-sample} with $\frac{1}{2 \lambda} \left[ \fMAP\left(\btheta + \bepsilon + \lambda \grad{\bz}f(\bz)\right) - \fMAP\left(\btheta + \bepsilon - \lambda \grad{\bz}f(\bz) \right) \right]$, leading to the update equation in \cref{algo:main}.
\section{Related Work}
\paragraph{Continuous relaxations.}
Several works address the gradient estimation problem for discrete random variables, often resorting to continuous relaxations.
\citet{maddison2016concrete,jang2016categorical} propose the Gumbel-Softmax (or concrete) distribution to relax categorical random variables, which was extended by \citet{paulus2020gradient} to more complex probability distributions.
The Gumbel-Softmax distribution only directly applies to categorical variables: for more complex distributions, one has to come up with tailor-made relaxations, or use the STE or SFE -- \eg, see \citet{kim2016exact} and \citet{grover2019stochastic}.
REBAR~\citep{tucker2017rebar} and RELAX~\citep{grathwohl2017backpropagation} use parameterized control variates based on continuous relaxations for the SFE.
In this work, we focus explicitly on problems where \emph{only} discrete samples are used during training.
Furthermore, REBAR is tailored to categorical distributions, while \imle and \aimle are intended for models with complex distributions and multiple constraints.
Approaches that do not rely on relaxations are specific to certain distributions~\citep{bengio2013estimating,franceschi2019learning,liu2019rao} or assume knowledge of the constraints $\mathcal{C}$~\citep{kool2020estimating}.
\aimle and \imle provide a general-purpose framework that does not require access to the linear constraints and the corresponding integer polytope $\mathcal{C}$.  
%
%
SparseMAP~\citep{Niculae2018SparseMAPDS} is an approach to structured prediction and latent variables, replacing an exponential distribution with a sparser distribution; similarly to our work, it only presupposes the availability of a MAP oracle.
LP-SparseMAP~\citep{Niculae2020LPSparseMAPDR} is an extension of SparseMAP that uses a relaxation of the underlying optimisation problem. 
\paragraph{Differentiating through combinatorial solvers.}
A series of works about differentiating through combinatorial optimisation problems~\citep{wilder2019melding,elmachtoub2020smart,ferber2020mipaal,DBLP:conf/nips/MandiG20} relax ILPs by adding a regularisation term, and differentiate through the KKT conditions deriving from the application of the cutting plane or the interior-point methods. 
These approaches are conceptually linked to techniques for differentiating through smooth programs~\citep{amos2017optnet,donti2017task,agrawal2019differentiable,chen2020understanding,domke2012generic,franceschi2018bilevel} that arise in modelling, hyperparameter optimisation, and meta-learning.
Black-box Backprop~\citep{poganvcic2019differentiation,rolinek2020deep} and DPO~\citep{berthet2020learning} are methods that are not tied to a specific ILP solver. 
%
%
Black-box Backprop, originally derived from a continuous interpolation argument, can be interpreted as special instantiations of \imle and \aimle.
DPO addresses the theory of perturbed optimizers and discusses Perturb-and-MAP in the context of the Fenchel-Young losses.
All the combinatorial optimisation-related works assume that either optimal costs or solutions are given as training data, while \imle and \aimle can also be applied in the absence of such supervision by making use of implicitly generated target distributions.
Other authors focus on devising differentiable relaxations for specific combinatorial problems such as SAT~\cite{evans2018learning} or MaxSAT~\cite{wang2019satnet}.
Machine learning intersects with combinatorial optimisation in other contexts, \eg in learning heuristics to improve the performances of combinatorial solvers --- we refer to \citet{bengio2020machine} for further details.

Direct Loss Minimisation~\citep[DLM,][]{mcallester2010direct,song2016training} is also related to our work, but it relies on the assumption that examples of optimal states $\bz$ are given.
\citet{lorberbom:2019} extend the DLM framework to discrete VAEs using coupled perturbations: their approach is tailored to VAEs, and is not general-purpose.
Under a methodological viewpoint, \imle inherits from classical MLE~\citep{wainwright2008graphical} and Perturb-and-MAP~\citep{Papandreou:2011}.
The theory of Perturb-and-MAP was used to derive general-purpose upper bounds for log-partition functions~\citep{hazan2012partition,shpakova:2016}.

\section{Experiments}
Similarly to \citet{niepert21imle}, conducted three different types of experiments.
First, we analyse and compare the behaviour of \aimle with other gradient estimators (STE, SFE, \imle) in a synthetic setting.
Second, we consider a setting where the distribution parameters $\btheta$ are produced by an upstream neural model, denoted by $f$ in \cref{eq:hybridm}, and the optimal discrete structure is not available during training.
Finally, we consider the problem of differentiating through black-box combinatorial solvers, where we use the target distribution derived in \cref{imle-main-equation}.
In all our experiments, we fix the \aimle hyper-parameters and use the target gradient norm $c$ to $c = 1$, and the update step $\eta$ to $\eta = 10^{-3}$, based on the \aimle implementation described in \cref{algo:main}.
More experimental details and additional experiments on the Warcraft dataset proposed by \citet{poganvcic2019differentiation} are available in the appendix. 

\begin{figure}
\centering
\includegraphics[width=\columnwidth]{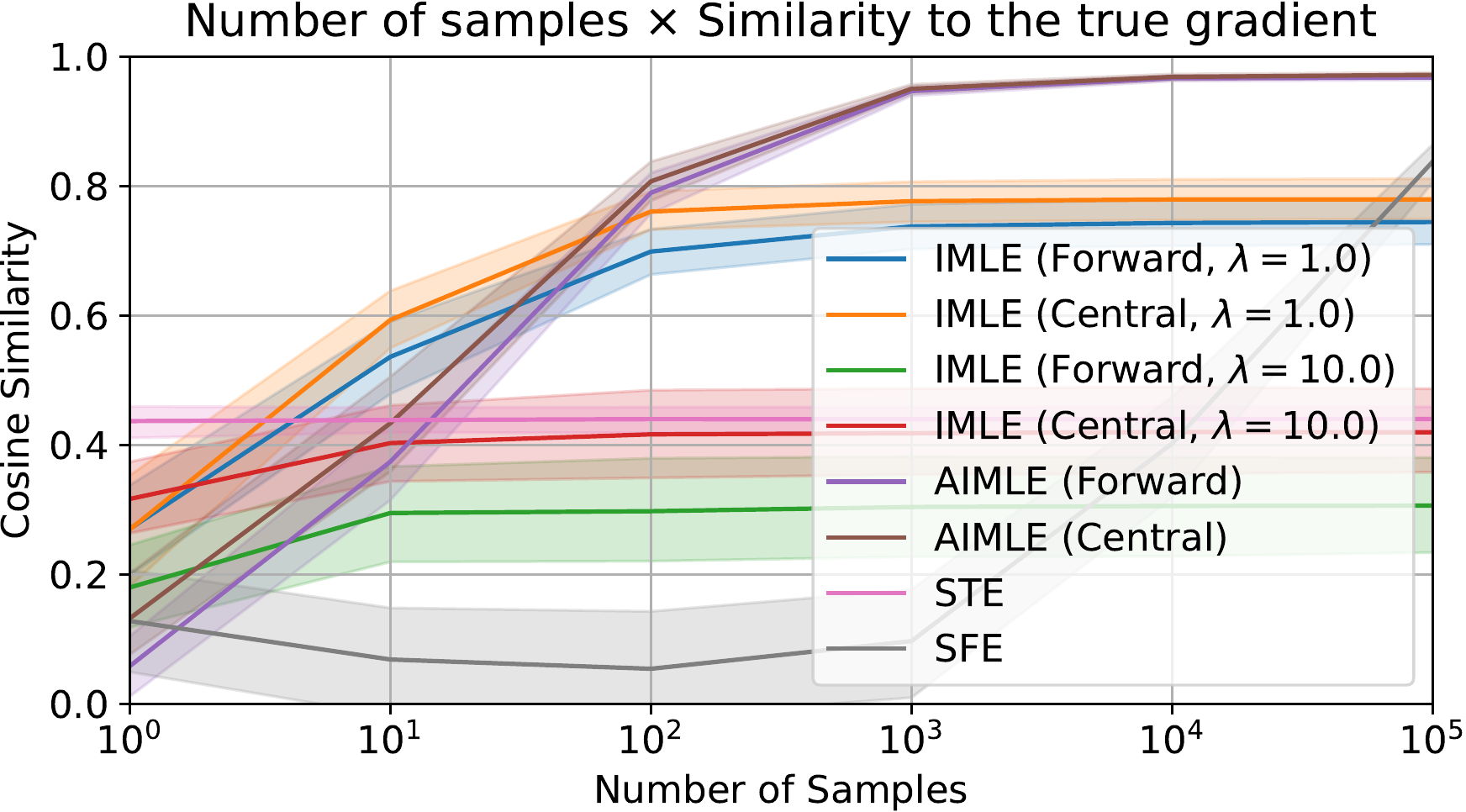}
%
%
\caption{Cosine similarity between the estimated gradient and the true gradient ($y$-axis) using several estimators (\imle, \aimle, STE, and SFE) with $S$ samples, with $S \in \{ 10^{0}, 10^{1}, \ldots, 10^{5} \}$ ($x$-axis) --- the gradient estimates produced by \aimle, both in its forward and central difference versions, are significantly more similar to the true gradient than the estimates produced by other methods.} \label{fig:estimators}
\end{figure}
\begin{figure}
\centering
\includegraphics[width=\columnwidth]{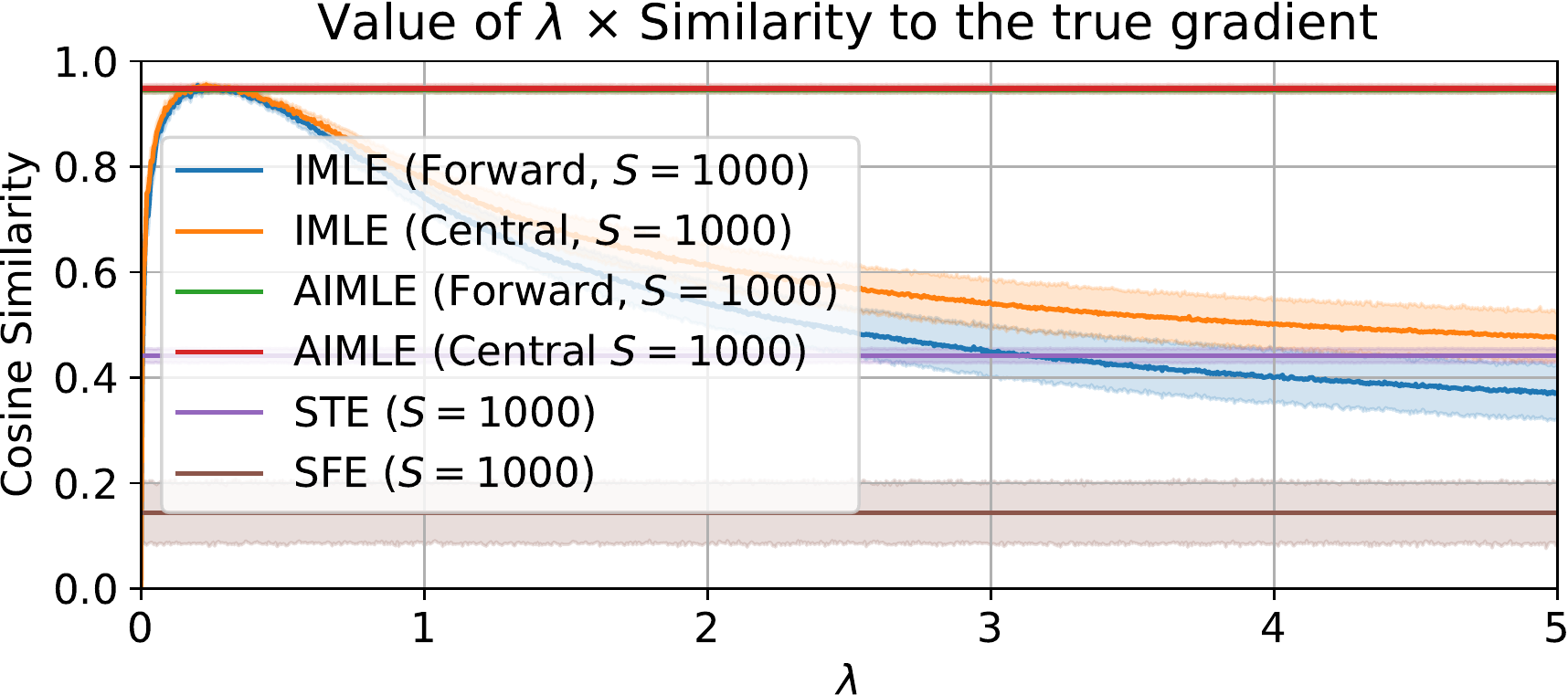}
\caption{Cosine similarity between the estimated gradient and the true gradient ($y$-axis) using several estimators --- namely \imle with a varying $\lambda \in [0, 5]$ ($x$-axis), \aimle, STE, and SFE --- with $S = 1000$ samples.} \label{fig:spot}
\end{figure}

\paragraph{Synthetic Experiments.}
We conducted a series of experiments with a tractable categorical distribution where $\bz \in \{0,1\}^n$ with $n \in \{ 10, 20, 30, 50 \}$ and $\sum_{i} \bz_{i} = 1$.
We set the loss to $L(\bm{\theta}) = \mathbb{E}_{\bz \sim p(\mathbf{z}; \bm{\theta})} [ \norm{\bz - \mathbf{b}}^2 ]$, where $\mathbf{b} \sim \mathcal{N}(0,\mathbf{I})$.
In \cref{fig:estimators}, we plot the cosine similarity between the gradient estimates produced by the Straight-Through Estimator~\citep[STE,][]{bengio2013estimating} the Score Function Estimator \citep[SFE,][]{DBLP:journals/ml/Williams92}, Implicit Maximum Likelihood Estimation~\citep[\imle,][]{niepert21imle}, and \aimle.
For STE and \imle, we use Perturb-and-MAP with Gumbel noise.
For \imle and \aimle, we evaluated both their \emph{forward difference} (Forward) and \emph{central difference} (Central) versions.
We evaluated all estimators using $S \in \{ 10^{0}, 10^{1}, \ldots, 10^{5} \}$ samples, and report how $S$ influences the cosine similarity between the gradient estimate and true gradient.
Statistics are over $32$ runs.
From the results, outlined in \cref{fig:estimators}, we can see that \aimle, both in its central and forward difference versions, produces significantly more accurate estimates of the true gradient compared to \imle, STE, and SFE, with orders of magnitude fewer samples.
Furthermore, we report the cosine similarity between the true and the estimated gradient for \aimle, STE, SFE, and \imle with a varying value of $\lambda \in [0, 5]$ --- all estimators use $S = 1000$ samples.
Results are outlined in \cref{fig:spot}: we can see that \aimle can produce gradient estimates that are comparable to the best estimates produced by \imle, without the need of training a $\lambda$ hyper-parameter.
\begin{table}
\resizebox{\columnwidth}{!}{
\begin{tabular}{rcccc}
\toprule
\multicolumn{1}{c}{\multirow{3}{*}{\bf Method}} & \multicolumn{2}{c}{\bf Test MSE}  & \multicolumn{2}{c}{\bf Subset Prec.}  \\
\cmidrule(lr){2-3} \cmidrule(lr){4-5}
& \multicolumn{1}{c}{\bf Mean} & \multicolumn{1}{c}{\bf SD} &  {\bf Mean} &  {\bf SD} \\

\midrule
\multicolumn{5}{c}{Aspect: \emph{aroma}, $K = 5$} \\
\midrule
SoftSub ($\tau=1.0$) &  2.515 &  0.087 &  55.453 &  2.338 \\
STE ($\tau=0.0$) &  4.660 &  0.053 &  44.593 &  0.523 \\
SST ($\tau=0.5$) &  4.788 &  0.486 &  56.854 &  3.752 \\
\imle (Forward, $\lambda=1000.0$, $\tau=1.0$) &  2.413 &  0.055 &  53.744 &  5.635 \\
\imle (Central, $\lambda=1000.0$, $\tau=0.0$) & \bf 2.266 & \bf 0.050 &  50.888 &  5.453 \\
\aimle (Forward, $\tau=1.0$) &  2.499 &  0.089 &  44.668 &  6.936 \\
\aimle (Central, $\tau=3.0$) &  2.385 &  0.049 & \bf 62.056 & \bf 2.107 \\
\midrule
\multicolumn{5}{c}{Aspect: \emph{aroma}, $K = 10$} \\
\midrule
SoftSub ($\tau=2.0$) &  2.543 &  0.044 &  44.513 &  2.958 \\
STE ($\tau=0.0$) &  4.310 &  0.039 &  39.635 &  0.281 \\
SST ($\tau=0.1$) &  5.213 &  0.295 &  24.328 &  12.463 \\
\imle (Forward, $\lambda=1000.0$, $\tau=1.0$) &  2.368 &  0.075 &  48.215 &  2.182 \\
\imle (Central, $\lambda=1000.0$, $\tau=0.0$) & \bf 2.256 & \bf 0.043 &  45.339 &  3.115 \\
\aimle (Forward, $\tau=2.0$) &  2.402 &  0.042 &  48.397 &  1.967 \\
\aimle (Central, $\tau=2.0$) &  2.419 &  0.061 & \bf 53.260 & \bf 2.271 \\
\midrule
\multicolumn{5}{c}{Aspect: \emph{aroma}, $K = 15$} \\
\midrule
SoftSub ($\tau=2.0$) &  2.711 &  0.035 &  37.202 &  1.374 \\
STE ($\tau=0.5$) &  4.062 &  0.054 &  36.267 &  0.161 \\
SST ($\tau=0.1$) &  5.787 &  0.517 &  24.551 &  9.827 \\
\imle (Forward, $\lambda=1000.0$, $\tau=1.0$) &  2.411 &  0.087 &  41.850 &  1.477 \\
\imle (Central, $\lambda=1000.0$, $\tau=0.0$) &  2.508 &  0.396 &  40.057 &  7.172 \\
\aimle (Forward, $\tau=1.0$) & \bf 2.408 & \bf 0.064 &  41.688 &  2.246 \\
\aimle (Central, $\tau=3.0$) &  2.470 &  0.026 & \bf 47.109 & \bf 2.863 \\

\bottomrule
\end{tabular}
}
\caption{Detailed results for the aspect \emph{aroma}. Test MSE and subset precision, both $\times 100$, for $k \in \{5, 10, 15\}$.} \label{tab:l2x-aroma}
\end{table}

\paragraph{Learning to Explain.} \label{section-experiments-l2x}

The \textsc{BeerAdvocate} dataset~\citep{McAuley2012LearningAA} consists of free-text reviews and ratings for $4$ different aspects of beer: \emph{appearance}, \emph{aroma}, \emph{palate}, and \emph{taste}.
Each sentence in the test set has annotations providing the words that best describe the various aspects.
Following the experimental setting in \citet{paulus2020gradient,niepert21imle}, we address the problem of learning a distribution over $k$-subsets of words that best explain a given aspect rating, introduced by \citet{chen2018learning}.
The complexity of the MAP problem for the $k$-subset distribution is linear in $k$.
The training set has 80,000 reviews for the aspect \emph{appearance} and 70,000 reviews for all other aspects. 
Since the original dataset~\citep{McAuley2012LearningAA} did not provide separate validation and test sets, following \citet{niepert21imle}, we compute 10 different evenly sized validation and test splits of the 10,000 held out set and compute mean and standard deviation over 10 models, each trained on one split.
Subset precision was computed using a subset of 993 annotated reviews.
We use pre-trained word embeddings from~\cite{Lei2016RationalizingNP}.
We extend the implementations provided by  \citet{niepert21imle}, which use a neural network following the architecture introduced by \citet{paulus2020gradient} with four convolutional layers and one dense layer.
This neural network outputs the parameters $\btheta$ of the distribution $p(\bz; \btheta)$ over $k$-hot binary latent masks with $k \in \{5, 10, 15\}$.
We compare \aimle (both the forward and central difference versions) to relaxation-based baselines L2X~\citep{chen2018learning} and SoftSub~\citep{Xie2019ReparameterizableSS}; to STE with Gumbel perturbations; and to \imle~\citep{niepert21imle} with Gumbel perturbations.
We used the standard hyperparameter settings of \citet{chen2018learning} and chose the temperature parameter $t \in \{0.1, 0.5, 1.0, 2.0\}$ for all methods.
For \imle we choose $\lambda \in \{10, 100, 1000\}$ based on the validation MSE.
We trained separate models for each aspect using MSE as the training loss, using the Adam~\citep{DBLP:journals/corr/KingmaB14} optimiser with its default hyper-parameters.
\cref{tab:l2x-aroma} lists detailed results for the aspect \emph{aroma}.
We can see that \aimle, in its central differences version, systematically produces the highest subset precision values while yielding test MSE values comparable to those produced by \imle, while not requiring tuning the $\lambda$ hyper-parameter.
In the appendix, we report the results for the other aspects, where we notice that \aimle produces significantly higher subset precision values in all other aspects.

\begin{figure}[t!]
\centering
  \includegraphics[width=1\columnwidth]{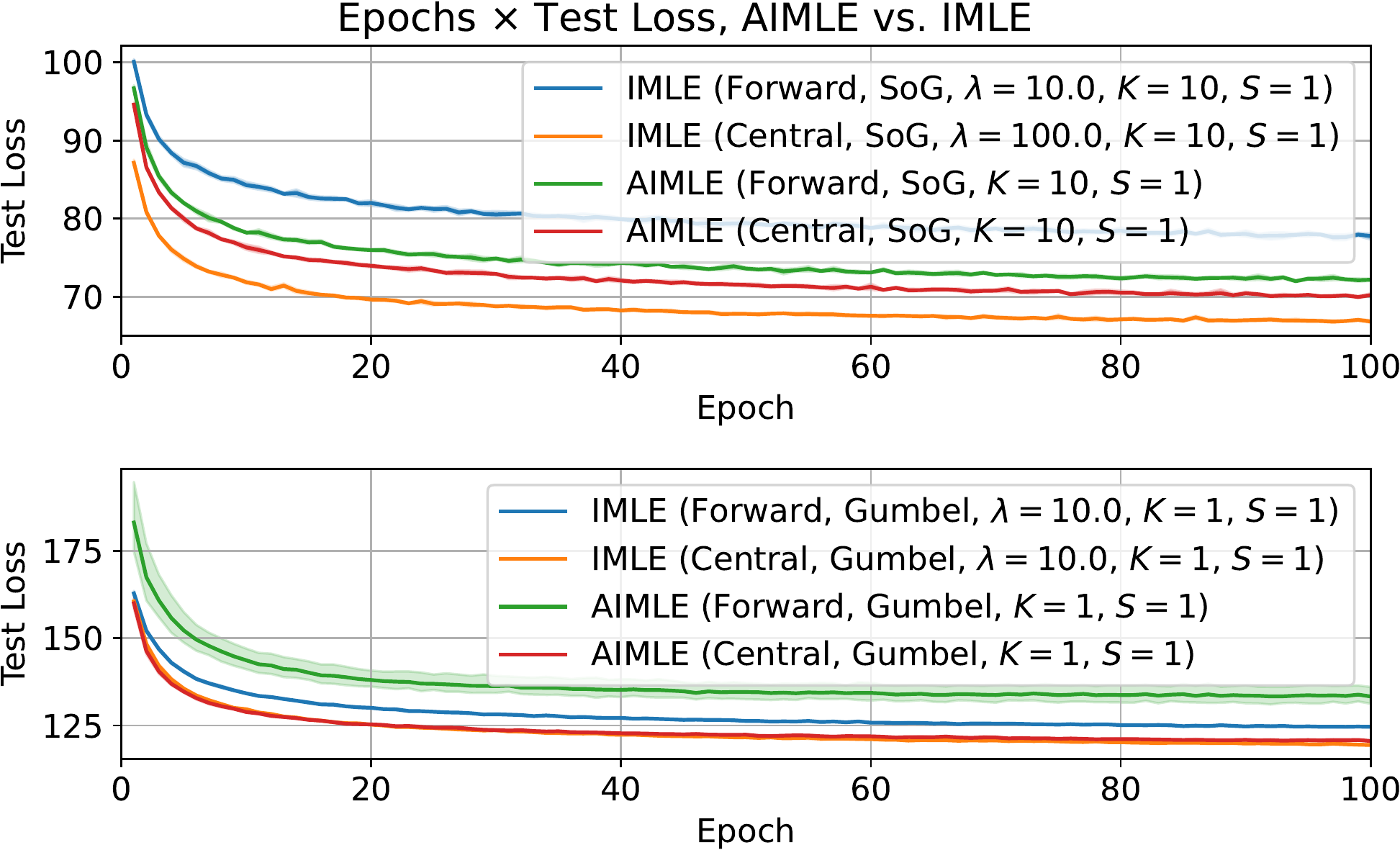}
\caption{Training dynamics for a DVAE using \aimle and \imle for $K=10$ (top) and $K=1$ (bottom).}
\label{fig:dvae-imle}
\end{figure}


\begin{figure}[t!]
\centering
  \includegraphics[width=1\columnwidth]{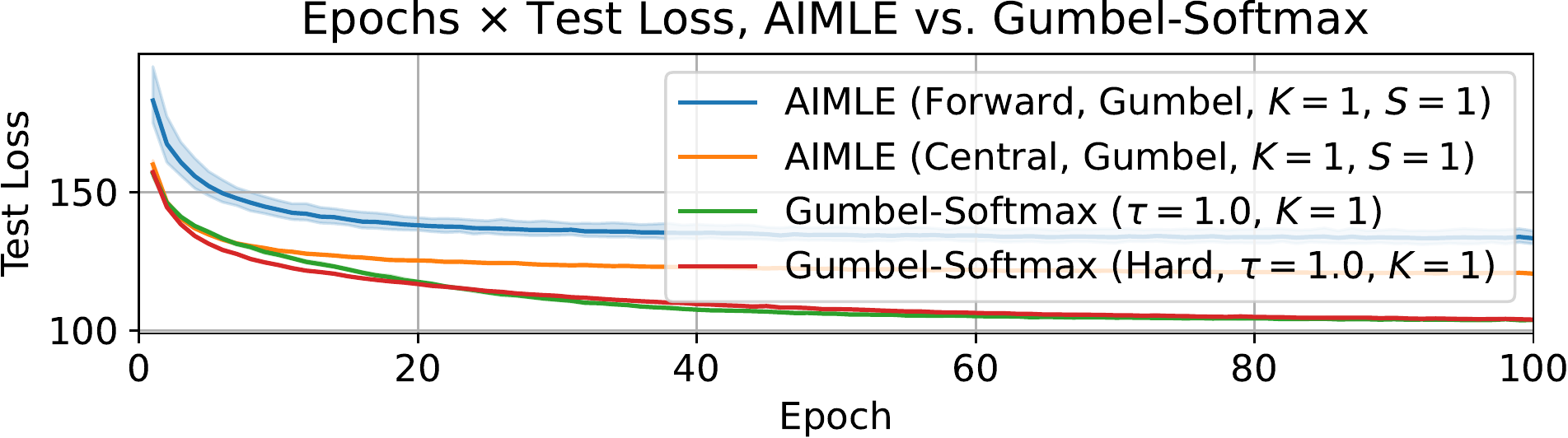}
\caption{Training dynamics for a DVAE using \aimle and Gumbel-Softmax for $K=1$.}
\label{fig:dvae-gs}
\end{figure}

\begin{figure}[t!]
\centering
  \includegraphics[width=1\columnwidth]{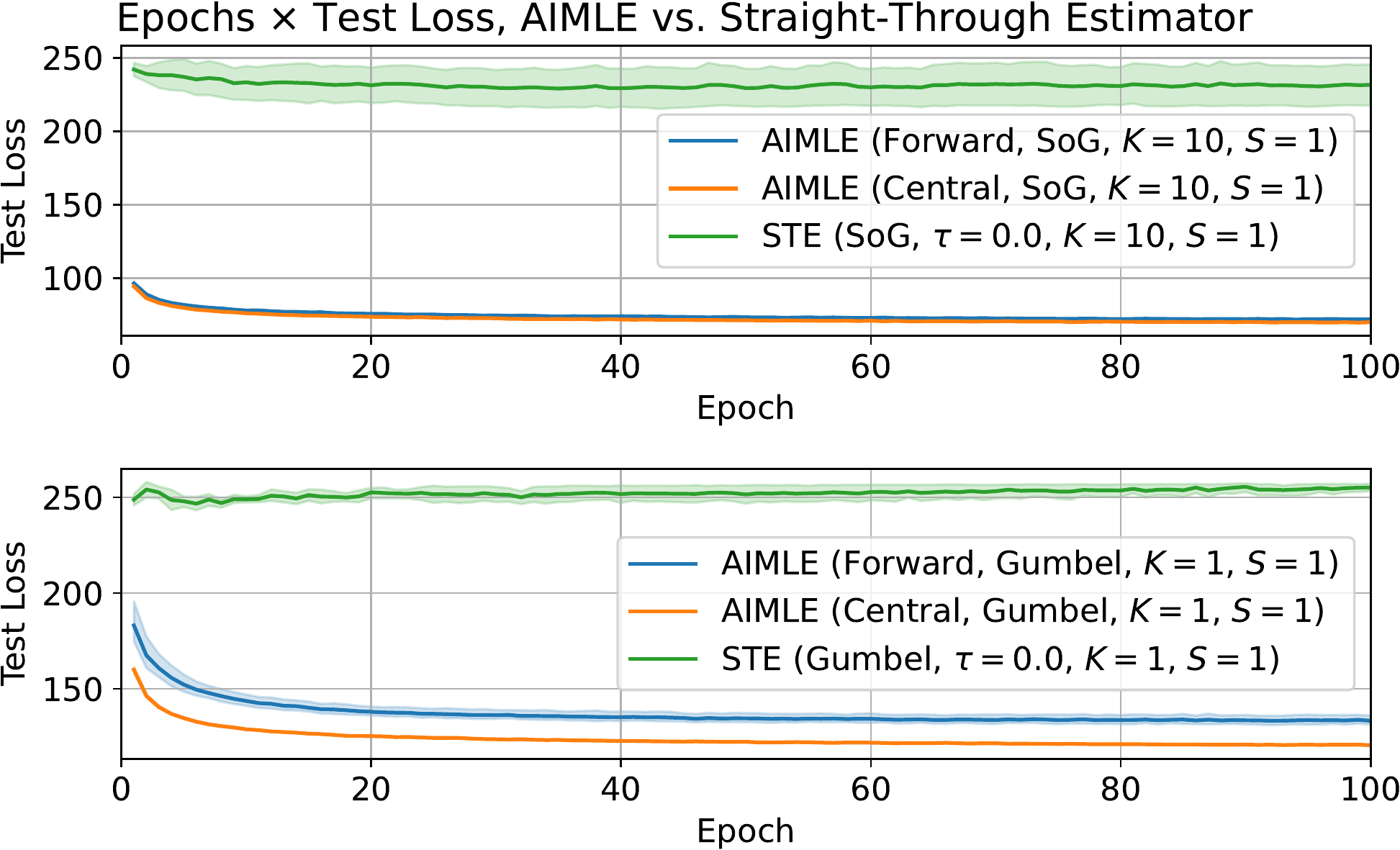}
\caption{Training dynamics for a DVAE using \aimle and STE for $K=10$ (top) and $K=1$ (bottom).}
\label{fig:dvae-ste}
\end{figure}

\begin{figure}[t!]
\centering
  \includegraphics[width=1\columnwidth]{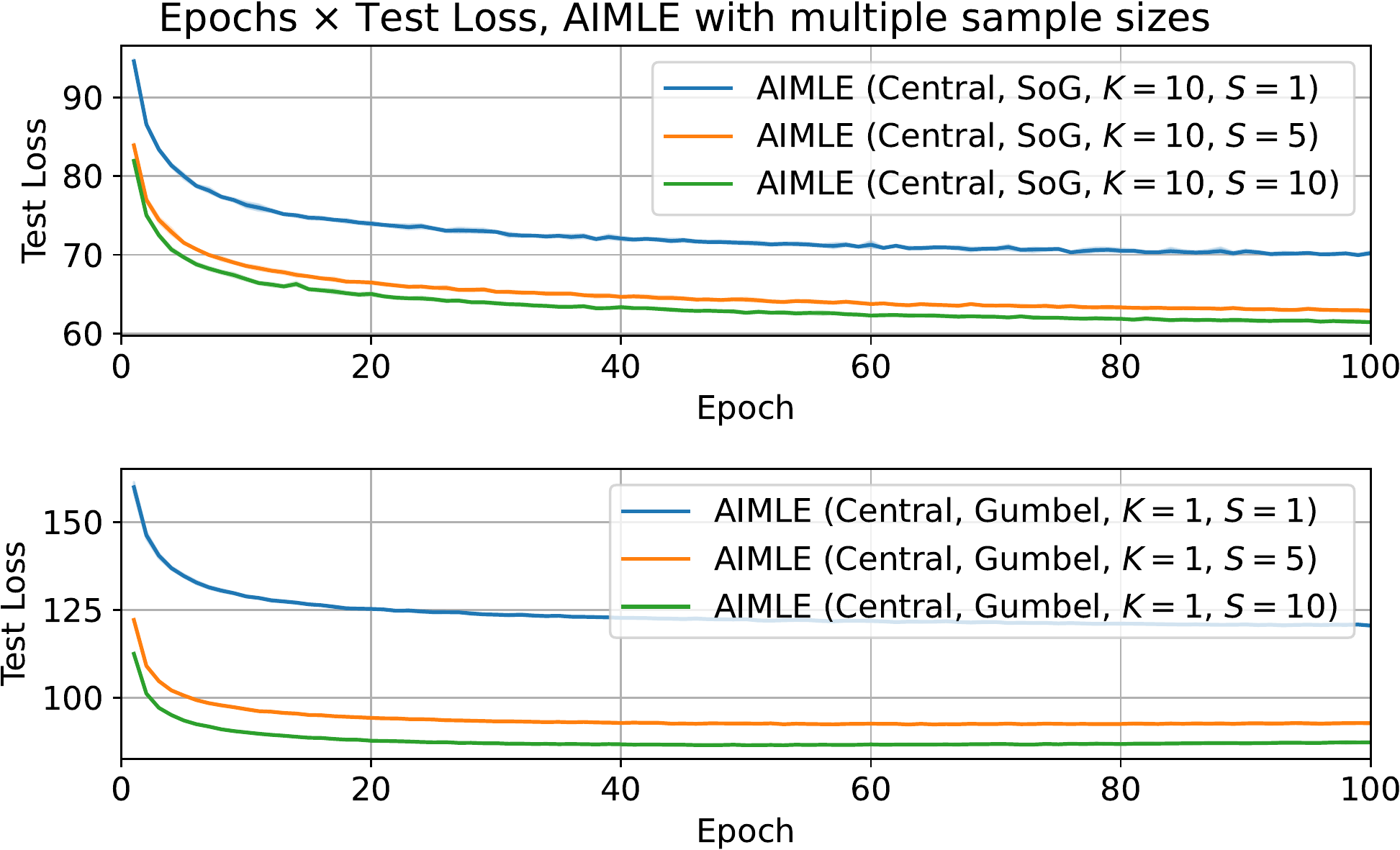}
\caption{Training dynamics for a  DVAE using \aimle with a different number of samples (\ie $S \in \{1, 5, 10 \}$) for $K=10$ (top) and $K=1$ (bottom).}
\label{fig:dvae-aimle}
\vspace{-15pt}
\end{figure}

%
%

\begin{table*}
  \caption{Latent Graph Structure Recovery -- Stochastic Softmax Tricks (SST,][]{paulus2020gradient} defining a \emph{spanning tree} over undirected edges (with hard sampling), in comparison with \imle~\citep{niepert21imle} and \aimle, where the MAP function is computed by Kruskal's algorithm~\citep{Kruskal1956}. \aimle yields the lowest test ELBO values, both in the T=10 (shorter sequences) and the T=20 (longer sequences) settings.}
  \label{table:graph_layout} 
  
  \resizebox{\linewidth}{!}{
  \begin{tabular}{rcccccc}
    \toprule
     & \multicolumn{3}{c}{T=10} & \multicolumn{3}{c}{T=20} \\ 
     \cmidrule(l){2-4} 
     \cmidrule(l){5-7}     
     Edge Distribution & ELBO & Edge Prec. & Edge Rec. & ELBO & Edge Prec. & Edge Rec. \\ 
    
    \midrule
    
    SST (Hard) & -2301.47 $\pm$ 85.86 & 33.75 $\pm$ 9.44 & 60.40 $\pm$ 23.23 & -3407.89 $\pm$ 221.53 & 57.40 $\pm$ 17.87 & 70.42 $\pm$ 8.22 \\
    
    \imle (Forward) & -2289.94 $\pm$ 4.31 & 23.94 $\pm$ 0.03 & 95.75 $\pm$ 0.14 & -3820.68 $\pm$ 25.32 & 20.28 $\pm$ 0.12 & 20.28 $\pm$ 0.12 \\
    
    \imle (Central) & -2341.71 $\pm$ 41.68 & 43.95 $\pm$ 7.22 & 43.95 $\pm$ 7.22 & -3447.29 $\pm$ 550.38 & 40.25 $\pm$ 14.26 & 40.25 $\pm$ 14.26 \\

    


    \aimle (Forward) & \bf -1877.90 $\pm$ 277.53 & 55.23 $\pm$ 11.86 & 55.23 $\pm$ 11.86 & \bf -1884.83 $\pm$ 124.62 & 40.48 $\pm$ 4.25 & 40.48 $\pm$ 4.25 \\

    \aimle (Central) & -2018.39 $\pm$ 357.16 & 29.32 $\pm$ 6.89 & 41.83 $\pm$ 21.51 & -1999.57 $\pm$ 856.27 & 70.89 $\pm$ 24.77 & 83.73 $\pm$ 1.31 \\
    
    \bottomrule
    \end{tabular}
}
\vspace{-10pt}
\end{table*}

\paragraph{Discrete Variational Auto-Encoder.}
Following \citet{niepert21imle}, we compare \imle, STE, and the Gumbel-Softmax trick~\citep{maddison2016concrete,jang2016categorical} using a discrete $K$-subset Variational Auto-Encoder (VAE).
The latent variables model a probability distribution over $K$-subsets of --- or top-$K$ assignments --- binary vectors of length 20; note that, for $K=1$, this is equivalent to a categorical variable with 20 categories.
We follow the implementation details in \citet{niepert21imle}, where the encoder and the decoder of the VAE consist of three dense layers, where the encoder and decoder activations have sizes 512-256-20$\times$20, and 256-512-784, respectively.
The loss is the sum of the reconstruction losses --- binary cross-entropy loss on output pixels --- and the KL divergence between the marginals of the variables and the uniform distribution.
For \aimle, \imle, and STE, we use $\GumbelDist(0, 1)$ perturbations for $K=1$, and Sum-of-Gamma perturbations~\citep[SoG,][]{niepert21imle} for $K=10$, with a temperature of $\tau = 1$.
For \imle, we select the hyper-parameter $\lambda \in \{ 1, 10, 100 \}$ on a held-out validation set based on the validation loss.
For \imle and \aimle, we report the results for both the forward and central difference versions.
We train the DVAE for $100$ epochs and report the loss on a held-out test set.
In \cref{fig:dvae-imle}, we show the test losses for \imle and \aimle (forward and central difference versions), for $K = 10$ and $K = 1$.
We can see that, for $K=10$, \aimle produces significantly lower test loss values than \imle, and, for $K=1$, the central difference version of \aimle produces test loss values comparable to \imle.
We also compare \aimle to the Gumbel-Softmax trick (see \cref{fig:dvae-gs}) and STE (see \cref{fig:dvae-ste}): \aimle produces significantly lower test losses than STE ($K \in \{ 1, 10 \}$) while producing higher test losses than Gumbel-Softmax ($K = 1$).
%
When increasing the number of samples $S$ in \aimle (see \cref{fig:dvae-aimle}), we found that higher values of $S$ produce significantly lower test loss values, to the point that, for $K = 1$, \aimle with $S \in \{ 5, 10 \}$ produces lower test loss values than Gumbel-Softmax, but with a higher computational cost.
All hyper-parameters and an analysis of the evolution of $\lambda$ during training are available in the appendix.
\paragraph{Neural Relational Inference for Recovering Latent Graph Structures.}
In this experiment, we investigated the use of \aimle for recovering latent graph structures and predicting the evolution of a dynamical system.
In Neural Relational Inference~\citep[NRI,][]{DBLP:conf/icml/KipfFWWZ18}, a Graph Neural Network~\citep[GNN,][]{DBLP:journals/tnn/Micheli09,DBLP:journals/tnn/ScarselliGTHM09} encoder is used to generate a latent interaction graph, which is then used to produce a distribution over an interacting particle system.
NRI is trained as a variational auto-encoder to maximise a lower bound (ELBO) on the marginal log-likelihood of the time series.
Based on the implementation provided by \citet{paulus2020gradient}, we compared the Stochastic Softmax Tricks~\citep[SST,][]{paulus2020gradient} encoder that induces a \emph{spanning tree} over undirected edges, with an encoder producing a maximum spanning tree using Kruskal's algorithm~\citep{Kruskal1956}, and using either \imle or \aimle to back-propagate through it.
In this setting, Kruskal's algorithm represents the MAP estimator for a distribution over the latent graph structures.
Our dataset consisted of latent prior spanning trees over 10 vertices sampled from the $\GumbelDist(0, 1)$ prior.
Given a tree, we embed the vertices in $\mathbb{R}^2$ by applying $T \in \{10, 20\}$ iterations of a force-directed algorithm~\citep{DBLP:journals/spe/FruchtermanR91}.
The model saw particle locations at each iteration, not the underlying spanning tree.
Results are outlined in \cref{table:graph_layout}: we found that \aimle performed best, improving on both ELBO and the recovery of latent structure over the structured SST baseline proposed by \citet{paulus2020gradient}.
\section{Conclusions}
We introduced Adaptive Implicit Maximum Likelihood Estimation (\aimle), an efficient, simple-to-implement, and general-purpose framework for learning hybrid models.
\aimle is an extension of \imle~\citep{niepert21imle} that, during training, can dynamically select the optimal target distribution by identifying the update step $\lambda$ that yields the desired gradient norm.
Furthermore, \aimle incorporates insights from finite difference methods, improving its effectiveness in gradient estimation tasks.
In our experiments, we show that \aimle produces better results than relaxation-based approaches for discrete latent variable models and approaches that back-propagate through black-box combinatorial solvers.
A limitation is that \aimle relies on a \emph{warm-up period} for selecting the optimal $\lambda$, whose duration varies depending on the update step $\eta$ --- which we fix to $\eta = 10^{-3}$.
A potential solution to this problem is to use adaptive update steps, such as momentum~\citep{DBLP:journals/nn/Qian99}.
\paragraph{Acknowledgements}
Pasquale was partially funded by the European Union’s Horizon 2020 research and innovation programme under grant agreement no. 875160, ELIAI (The Edinburgh Laboratory for Integrated Artificial Intelligence) EPSRC (grant no. EP/W002876/1), an industry grant from Cisco, and a donation from Accenture LLP; and is grateful to NVIDIA for the GPU donations.
Mathias was partially funded by Deutsche Forschungsgemeinschaft (DFG, German Research Foundation) under Germany’s Excellence Strategy - EXC 2075.

\bibliography{ge}

\ifbool{printApdx}{\appendix

\section{Learning to Explain Experiments} \label{app:l2x}

\begin{table}
\resizebox{\columnwidth}{!}{
\begin{tabular}{rcccc}
\toprule
\multicolumn{1}{c}{\multirow{3}{*}{\bf Method}} & \multicolumn{2}{c}{\bf Test MSE}  & \multicolumn{2}{c}{\bf Subset Prec.}  \\
\cmidrule(lr){2-3} \cmidrule(lr){4-5}
& \multicolumn{1}{c}{\bf Mean} & \multicolumn{1}{c}{\bf SD} &  {\bf Mean} &  {\bf SD} \\

\midrule
\multicolumn{5}{c}{Aspect: \emph{appearance}, $K = 5$} \\
\midrule
SoftSub ($\tau=1.0$) &  2.327 &  0.051 &  72.183 &  5.242 \\
STE ($\tau=0.0$) &  5.315 &  0.070 &  29.134 &  1.503 \\
SST ($\tau=0.5$) &  4.157 &  0.562 &  76.059 &  7.957 \\
\imle (Forward, $\lambda=1000.0$, $\tau=1.0$) &  2.420 &  0.066 &  54.457 &  7.165 \\
\imle (Central, $\lambda=1000.0$, $\tau=0.0$) & \bf 2.263 & \bf 0.086 &  69.679 &  12.590 \\
\aimle (Forward, $\tau=3.0$) &  2.512 &  0.058 &  56.904 &  4.149 \\
\aimle (Central, $\tau=3.0$) &  2.266 &  0.071 & \bf 81.119 & \bf 2.345 \\
\midrule
\multicolumn{5}{c}{Aspect: \emph{appearance}, $K = 10$} \\
\midrule
SoftSub ($\tau=2.0$) &  2.349 &  0.064 &  63.667 &  5.215 \\
STE ($\tau=2.0$) &  4.967 &  0.049 &  33.629 &  2.268 \\
SST ($\tau=0.5$) &  4.973 &  0.443 &  60.363 &  12.225 \\
\imle (Forward, $\lambda=1000.0$, $\tau=1.0$) &  2.300 &  0.071 &  54.667 &  6.989 \\
\imle (Central, $\lambda=1000.0$, $\tau=0.0$) & \bf 2.168 & \bf 0.051 &  63.249 &  6.251 \\
\aimle (Forward, $\tau=1.0$) &  2.305 &  0.065 &  59.840 &  4.859 \\
\aimle (Central, $\tau=1.0$) &  2.191 &  0.037 & \bf 72.167 & \bf 2.744 \\
\midrule
\multicolumn{5}{c}{Aspect: \emph{appearance}, $K = 15$} \\
\midrule
SoftSub ($\tau=2.0$) &  2.534 &  0.132 &  54.097 &  7.032 \\
STE ($\tau=1.5$) &  4.493 &  0.083 &  37.737 &  3.298 \\
SST ($\tau=0.1$) &  5.537 &  0.605 &  36.217 &  18.978 \\
\imle (Forward, $\lambda=1000.0$, $\tau=1.0$) &  2.245 &  0.059 &  52.654 &  3.913 \\
\imle (Central, $\lambda=1000.0$, $\tau=0.0$) & \bf 2.192 & \bf 0.047 &  51.246 &  9.288 \\
\aimle (Forward, $\tau=3.0$) &  2.282 &  0.068 &  52.188 &  4.306 \\
\aimle (Central, $\tau=3.0$) &  2.195 &  0.046 & \bf 68.155 & \bf 3.206 \\

\bottomrule
\end{tabular}
}
\caption{Detailed results for the aspect \emph{appearance}. Test MSE and subset precision, both $\times 100$, for $k \in \{5, 10, 15\}$.} \label{app:l2x-appearance}
\end{table}

\begin{table}
\resizebox{\columnwidth}{!}{
\begin{tabular}{rcccc}
\toprule
\multicolumn{1}{c}{\multirow{3}{*}{\bf Method}} & \multicolumn{2}{c}{\bf Test MSE}  & \multicolumn{2}{c}{\bf Subset Prec.}  \\
\cmidrule(lr){2-3} \cmidrule(lr){4-5}
& \multicolumn{1}{c}{\bf Mean} & \multicolumn{1}{c}{\bf SD} &  {\bf Mean} &  {\bf SD} \\

\midrule
\multicolumn{5}{c}{Aspect: \emph{aroma}, $K = 5$} \\
\midrule
SoftSub ($\tau=1.0$) &  2.515 &  0.087 &  55.453 &  2.338 \\
STE ($\tau=0.0$) &  4.660 &  0.053 &  44.593 &  0.523 \\
SST ($\tau=0.5$) &  4.788 &  0.486 &  56.854 &  3.752 \\
\imle (Forward, $\lambda=1000.0$, $\tau=1.0$) &  2.413 &  0.055 &  53.744 &  5.635 \\
\imle (Central, $\lambda=1000.0$, $\tau=0.0$) & \bf 2.266 & \bf 0.050 &  50.888 &  5.453 \\
\aimle (Forward, $\tau=1.0$) &  2.499 &  0.089 &  44.668 &  6.936 \\
\aimle (Central, $\tau=3.0$) &  2.385 &  0.049 & \bf 62.056 & \bf 2.107 \\
\midrule
\multicolumn{5}{c}{Aspect: \emph{aroma}, $K = 10$} \\
\midrule
SoftSub ($\tau=2.0$) &  2.543 &  0.044 &  44.513 &  2.958 \\
STE ($\tau=0.0$) &  4.310 &  0.039 &  39.635 &  0.281 \\
SST ($\tau=0.1$) &  5.213 &  0.295 &  24.328 &  12.463 \\
\imle (Forward, $\lambda=1000.0$, $\tau=1.0$) &  2.368 &  0.075 &  48.215 &  2.182 \\
\imle (Central, $\lambda=1000.0$, $\tau=0.0$) & \bf 2.256 & \bf 0.043 &  45.339 &  3.115 \\
\aimle (Forward, $\tau=2.0$) &  2.402 &  0.042 &  48.397 &  1.967 \\
\aimle (Central, $\tau=2.0$) &  2.419 &  0.061 & \bf 53.260 & \bf 2.271 \\
\midrule
\multicolumn{5}{c}{Aspect: \emph{aroma}, $K = 15$} \\
\midrule
SoftSub ($\tau=2.0$) &  2.711 &  0.035 &  37.202 &  1.374 \\
STE ($\tau=0.5$) &  4.062 &  0.054 &  36.267 &  0.161 \\
SST ($\tau=0.1$) &  5.787 &  0.517 &  24.551 &  9.827 \\
\imle (Forward, $\lambda=1000.0$, $\tau=1.0$) &  2.411 &  0.087 &  41.850 &  1.477 \\
\imle (Central, $\lambda=1000.0$, $\tau=0.0$) &  2.508 &  0.396 &  40.057 &  7.172 \\
\aimle (Forward, $\tau=1.0$) & \bf 2.408 & \bf 0.064 &  41.688 &  2.246 \\
\aimle (Central, $\tau=3.0$) &  2.470 &  0.026 & \bf 47.109 & \bf 2.863 \\

\bottomrule
\end{tabular}
}
\caption{Detailed results for the aspect \emph{aroma}. Test MSE and subset precision, both $\times 100$, for $k \in \{5, 10, 15\}$.} \label{app:l2x-aroma}
\end{table}

\begin{table}
\resizebox{\columnwidth}{!}{
\begin{tabular}{rcccc}
\toprule
\multicolumn{1}{c}{\multirow{3}{*}{\bf Method}} & \multicolumn{2}{c}{\bf Test MSE}  & \multicolumn{2}{c}{\bf Subset Prec.}  \\
\cmidrule(lr){2-3} \cmidrule(lr){4-5}
& \multicolumn{1}{c}{\bf Mean} & \multicolumn{1}{c}{\bf SD} &  {\bf Mean} &  {\bf SD} \\

\midrule
\multicolumn{5}{c}{Aspect: \emph{palate}, $K = 5$} \\
\midrule
SoftSub ($\tau=2.0$) &  2.857 &  0.049 &  51.435 &  2.258 \\
STE ($\tau=0.5$) &  4.456 &  0.038 &  30.650 &  0.449 \\
SST ($\tau=0.5$) &  4.357 &  0.596 &  50.634 &  4.587 \\
\imle (Forward, $\lambda=1000.0$, $\tau=1.0$) &  2.867 &  0.042 &  50.066 &  1.262 \\
\imle (Central, $\lambda=1000.0$, $\tau=0.0$) &  2.684 &  0.047 &  54.347 &  1.320 \\
\aimle (Forward, $\tau=1.0$) &  2.856 &  0.050 &  47.818 &  4.231 \\
\aimle (Central, $\tau=1.0$) & \bf 2.683 & \bf 0.037 & \bf 56.043 & \bf 1.540 \\
\midrule
\multicolumn{5}{c}{Aspect: \emph{palate}, $K = 10$} \\
\midrule
SoftSub ($\tau=2.0$) &  2.957 &  0.048 &  35.604 &  1.818 \\
STE ($\tau=0.0$) &  4.065 &  0.039 &  32.096 &  0.501 \\
SST ($\tau=0.5$) &  4.754 &  0.260 &  37.302 &  7.304 \\
\imle (Forward, $\lambda=1000.0$, $\tau=1.0$) &  2.833 &  0.039 &  43.305 &  2.993 \\
\imle (Central, $\lambda=1000.0$, $\tau=0.0$) &  2.669 &  0.046 &  45.258 &  2.335 \\
\aimle (Forward, $\tau=2.0$) &  2.837 &  0.062 &  40.777 &  3.054 \\
\aimle (Central, $\tau=1.0$) & \bf 2.666 & \bf 0.020 & \bf 49.895 & \bf 3.750 \\
\midrule
\multicolumn{5}{c}{Aspect: \emph{palate}, $K = 15$} \\
\midrule
SoftSub ($\tau=2.0$) &  3.138 &  0.051 &  27.927 &  1.651 \\
STE ($\tau=0.0$) &  3.849 &  0.047 &  30.727 &  0.947 \\
SST ($\tau=0.5$) &  5.171 &  0.645 &  24.095 &  9.535 \\
\imle (Forward, $\lambda=1000.0$, $\tau=1.0$) &  2.892 &  0.058 &  35.700 &  2.847 \\
\imle (Central, $\lambda=1000.0$, $\tau=0.0$) &  2.808 &  0.210 &  36.916 &  6.323 \\
\aimle (Forward, $\tau=2.0$) &  2.885 &  0.027 &  36.070 &  2.279 \\
\aimle (Central, $\tau=2.0$) & \bf 2.708 & \bf 0.031 & \bf 44.860 & \bf 1.549 \\

\bottomrule
\end{tabular}
}
\caption{Detailed results for the aspect \emph{palate}. Test MSE and subset precision, both $\times 100$, for $k \in \{5, 10, 15\}$.} \label{app:l2x-palate}
\end{table}

\begin{table}
\resizebox{\columnwidth}{!}{
\begin{tabular}{rcccc}
\toprule
\multicolumn{1}{c}{\multirow{3}{*}{\bf Method}} & \multicolumn{2}{c}{\bf Test MSE}  & \multicolumn{2}{c}{\bf Subset Prec.}  \\
\cmidrule(lr){2-3} \cmidrule(lr){4-5}
& \multicolumn{1}{c}{\bf Mean} & \multicolumn{1}{c}{\bf SD} &  {\bf Mean} &  {\bf SD} \\

\midrule
\multicolumn{5}{c}{Aspect: \emph{taste}, $K = 5$} \\
\midrule
SoftSub ($\tau=1.0$) &  2.196 &  0.045 &  42.762 &  1.785 \\
STE ($\tau=0.0$) &  4.591 &  0.047 &  39.360 &  0.402 \\
SST ($\tau=0.5$) &  3.942 &  0.354 &  35.376 &  4.146 \\
\imle (Forward, $\lambda=1000.0$, $\tau=1.0$) &  2.253 &  0.036 &  39.846 &  1.855 \\
\imle (Central, $\lambda=1000.0$, $\tau=0.0$) & \bf 2.124 & \bf 0.060 &  41.105 &  1.054 \\
\aimle (Forward, $\tau=1.0$) &  2.196 &  0.054 &  38.344 &  2.007 \\
\aimle (Central, $\tau=3.0$) &  2.132 &  0.043 & \bf 43.539 & \bf 1.175 \\
\midrule
\multicolumn{5}{c}{Aspect: \emph{taste}, $K = 10$} \\
\midrule
SoftSub ($\tau=2.0$) &  2.137 &  0.048 &  42.716 &  1.085 \\
STE ($\tau=0.0$) &  4.283 &  0.053 &  38.089 &  0.089 \\
SST ($\tau=0.1$) &  4.422 &  0.176 &  30.581 &  4.077 \\
\imle (Forward, $\lambda=1000.0$, $\tau=1.0$) &  2.200 &  0.052 &  40.673 &  1.756 \\
\imle (Central, $\lambda=1000.0$, $\tau=0.0$) & \bf 2.081 & \bf 0.054 &  39.636 &  1.219 \\
\aimle (Forward, $\tau=2.0$) &  2.190 &  0.030 &  41.098 &  1.525 \\
\aimle (Central, $\tau=1.0$) &  2.111 &  0.034 & \bf 45.593 & \bf 1.222 \\
\midrule
\multicolumn{5}{c}{Aspect: \emph{taste}, $K = 15$} \\
\midrule
SoftSub ($\tau=2.0$) &  2.173 &  0.043 &  40.189 &  1.318 \\
STE ($\tau=0.0$) &  4.009 &  0.035 &  37.364 &  0.058 \\
SST ($\tau=0.5$) &  4.868 &  0.264 &  32.507 &  3.649 \\
\imle (Forward, $\lambda=1000.0$, $\tau=1.0$) &  2.201 &  0.052 &  40.636 &  1.479 \\
\imle (Central, $\lambda=1000.0$, $\tau=0.0$) &  2.193 &  0.270 &  38.987 &  1.680 \\
\aimle (Forward, $\tau=3.0$) &  2.196 &  0.046 &  40.993 &  1.627 \\
\aimle (Central, $\tau=3.0$) & \bf 2.138 & \bf 0.041 & \bf 43.089 & \bf 1.780 \\

\bottomrule
\end{tabular}
}
\caption{Detailed results for the aspect \emph{taste}. Test MSE and subset precision, both $\times 100$, for $k \in \{5, 10, 15\}$.} \label{app:l2x-taste}
\end{table}

Here we report the results on all aspects of the Learning to Explain task with the \textsc{BeerAdvocate} dataset~\citep{McAuley2012LearningAA}, namely \emph{appearance} (\cref{app:l2x-appearance}), \emph{aroma} (\cref{app:l2x-aroma}), \emph{palate} (\cref{app:l2x-palate}) and \emph{taste} (\cref{app:l2x-taste}).
In all cases, \aimle produces the best subset precision results while producing test MSE values comparable with those produced by \imle.
Each experiment was re-run ten times for 20 training epochs; batch size was set to 40, with a kernel size of 3, hidden dimension of 250, and maximum sequence length of 350.
All models were trained using the Adam~\citep{DBLP:journals/corr/KingmaB14} optimiser, with a learning rate of $10^{-3}$, by fitting an MSE loss between the predicted and true ratings.
For all methods, the noise temperature $\tau$ was selected in $\tau \in \{ 0, 1, 2, 3 \}$.
In \imle, the hyper-parameter $\lambda$ was selected in $\{ 10, 100, 1000 \}$.
In \aimle, the hyper-parameters $c$ (the target gradient norm) and $\eta$ (the $\lambda$ update step) were fixed to their default values, namely $c = 1$ and $\eta = 10^{-3}$.

\section{Discrete Variational Auto-Encoder Experiments}

The data set can be loaded in PyTorch with \textit{torchvision.datasets.MNIST}.
As in prior work, we use a  batch size of $100$ and train for $100$ epochs, plotting the test loss after each epoch. We use the standard Adam settings in PyTorch and no learning rate schedule.
The MNIST dataset consists of black-and-white $28 \times 28$ pixels images of hand-written digits.
The encoder network consists of an input layer with dimension $784$ since we flatten the images; a dense layer with dimension $512$ and ReLU~\citep{DBLP:conf/icml/NairH10} activations; a dense layer with dimension $256$ and ReLU activations; and a dense layer with dimension $400$ ($20 \times 20$) which outputs the $\btheta$ and no non-linearity.
The layer implementing \aimle receives $\btheta$ as input and outputs a discrete latent code with shape $20\times20$.
The decoder, which takes this discrete latent code as input, consists of a dense layer with dimension $256$ and ReLU activation; a dense layer with dimension $512$ and ReLU activations; and finally, a dense layer with dimension $784$ returning the logits for the output pixels.
Sigmoid non-linearities are applied to these logits and used to compute the binary cross-entropy.

\begin{figure}[t!]
  \centering
    \includegraphics[width=1\columnwidth]{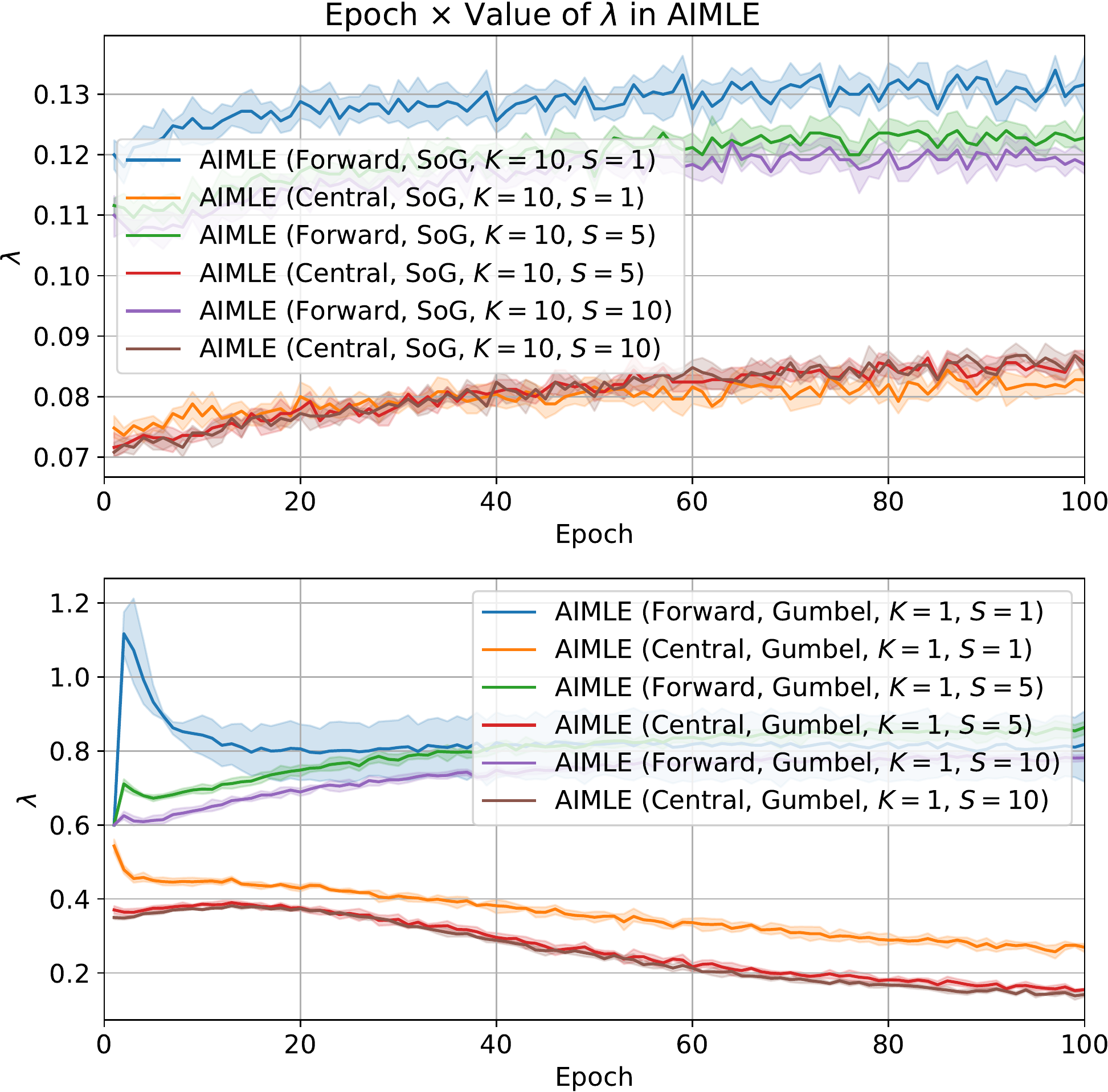}
  \caption{Training dynamics for a  DVAE using \aimle with a different number of samples (\ie $S \in \{1, 5, 10 \}$) for $K=10$ (top) and $K=1$ (bottom).}
  \label{fig:dvae-aimle-dyn}
\end{figure}

\paragraph{Evolution of $\lambda$ during training.}
In \cref{fig:dvae-aimle-dyn}, we show how $\lambda$ in AIMLE evolves during training.
We can see that the $\lambda$ values inferred by the adaptive method were lower for the central difference variant of AIMLE and that $K = 1$ results in higher inferred values for $\lambda$ compared to $K = 10$, which can be explained by the fact that the former requires higher values of $\lambda$ for altering the output of the $\fMAP$ (top-$K$) function for obtaining non-zero gradient estimates.

\section{Neural Relational Inference Experiments}
We use the dataset of latent prior spanning trees over ten vertices proposed by \citet{paulus2020gradient}: latent spanning trees were sampled by applying Kruskal's algorithm~\citep{Kruskal1956} to $U \sim \GumbelDist(0, 1)$ for a fully-connected graph.
Initial vertex locations were sampled from $\mathcal{N}(0, I)$ in $\mathbb{R}^{2}$.
Given the initial locations and the latent tree, the dynamical observations were obtained by applying a force-directed algorithm for graph layout for $T \in \{ 10, 20 \}$ iterations.
Then, the initial vertex positions were discarded because the first iteration of the layout algorithm typically results in large relocations.
Hence, the final dataset used for training consisted of 10 and 20 location observations in $\mathbb{R}^{2}$ for each of the 10 vertices.
Following this procedure, we generated a training set of size 50,000 and validation and test sets of size 10,000.
\paragraph{Model}
We follow the design and implementation provided by \citet{paulus2020gradient}, where the NRI model consists of two neural modules, an \emph{encoder} and a \emph{decoder}.
The encoder GNN passes messages over the fully connected directed graph with $n = 10$ nodes.
We took the final edge representations produced by the GNN and used them as $\btheta$.
The final edge representations are in $\mathbb{R}^{90 \times m}$, where $m = 1$ over the 90 undirected edges (since we do not consider self-connections).
Given the previous time-step data, the decoder GNN passes messages over the sampled graph adjacency matrix $X$ and predicts future node positions.
As in \citet{DBLP:conf/icml/KipfFWWZ18} and \citet{paulus2020gradient}, we used teacher forcing every ten time-steps.
$X \in \{ 0, 1 \}^{n \times n}$, in this case, was a directed adjacency matrix over the graph $G = (V, E)$ where $V$ are the nodes: $X_{ij} = 1$ is interpreted as there being an edge from $v_{i} \in V$ to $v_{j} \in V$, and $X_{ij} = 0$ otherwise.
$X$ represents the symmetric, directed adjacency matrix with edges in both directions for each undirected edge.
The decoder passes messages between both connected and not-connected nodes.
When considering a message from $v_{i}$ to $v_{j}$, it uses one network for the edges such that $X_{ij} = 1$, and another network for the edges such that $X_{ij} = 0$.
\begin{figure}[t]
    \centering
      \includegraphics[width=1\columnwidth]{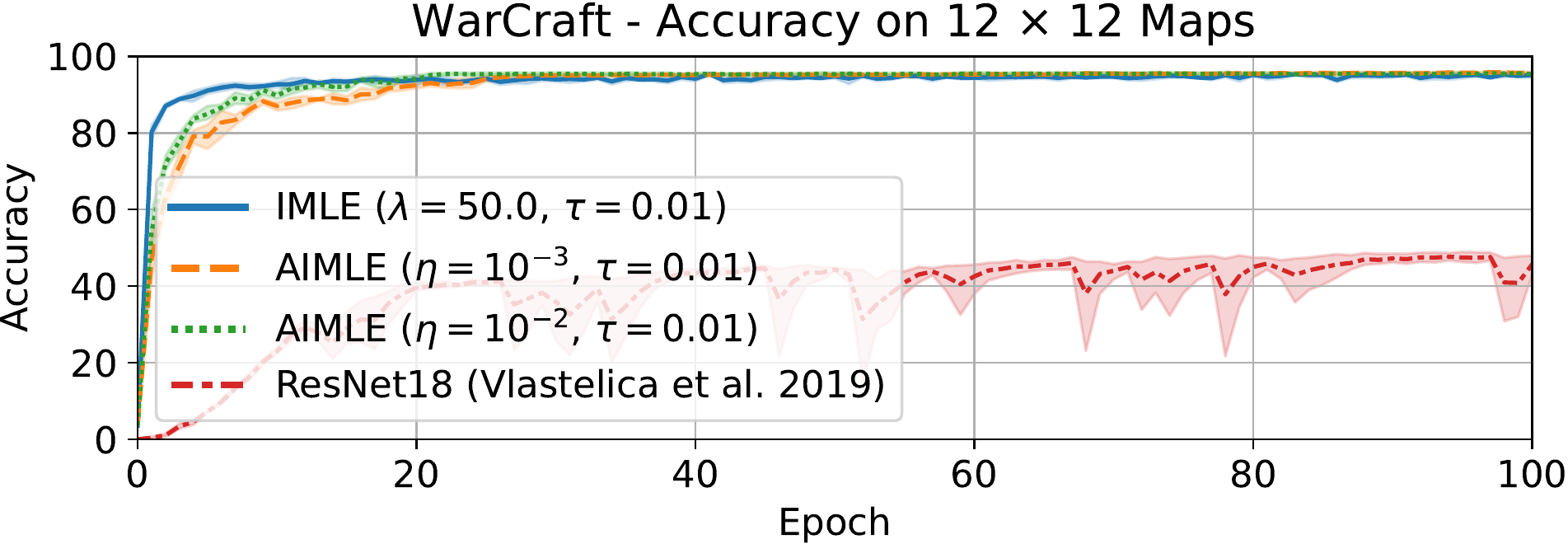}
      \includegraphics[width=1\columnwidth]{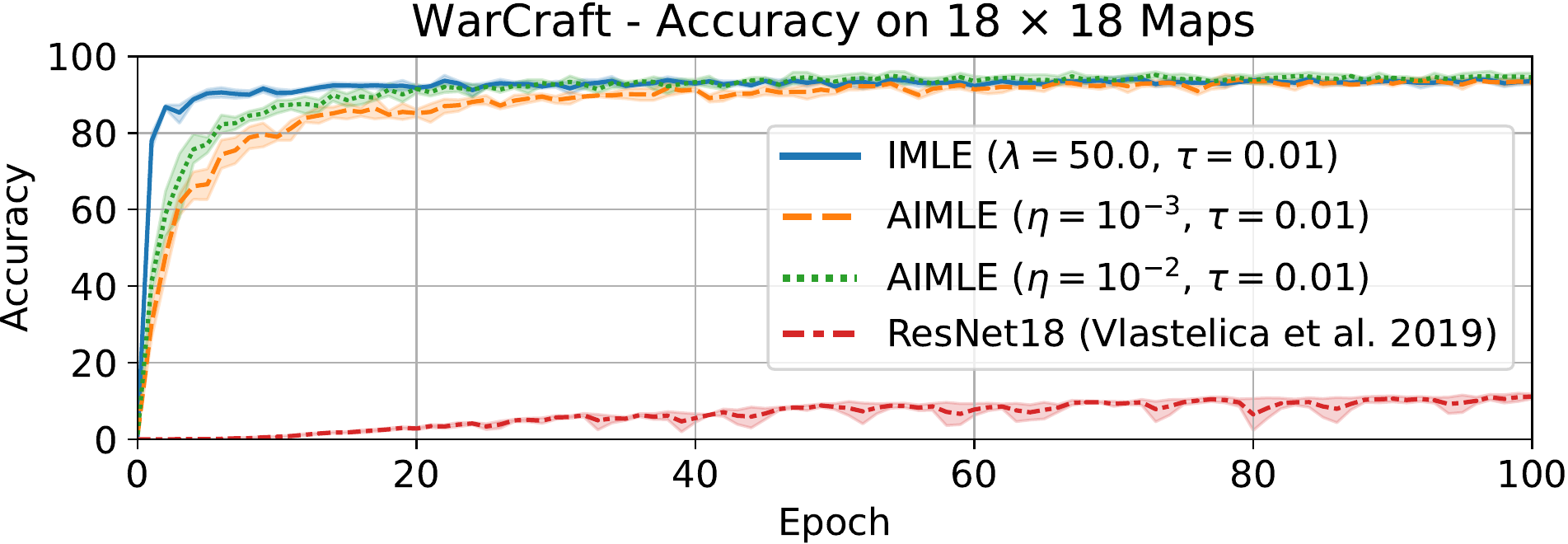}
      \includegraphics[width=1\columnwidth]{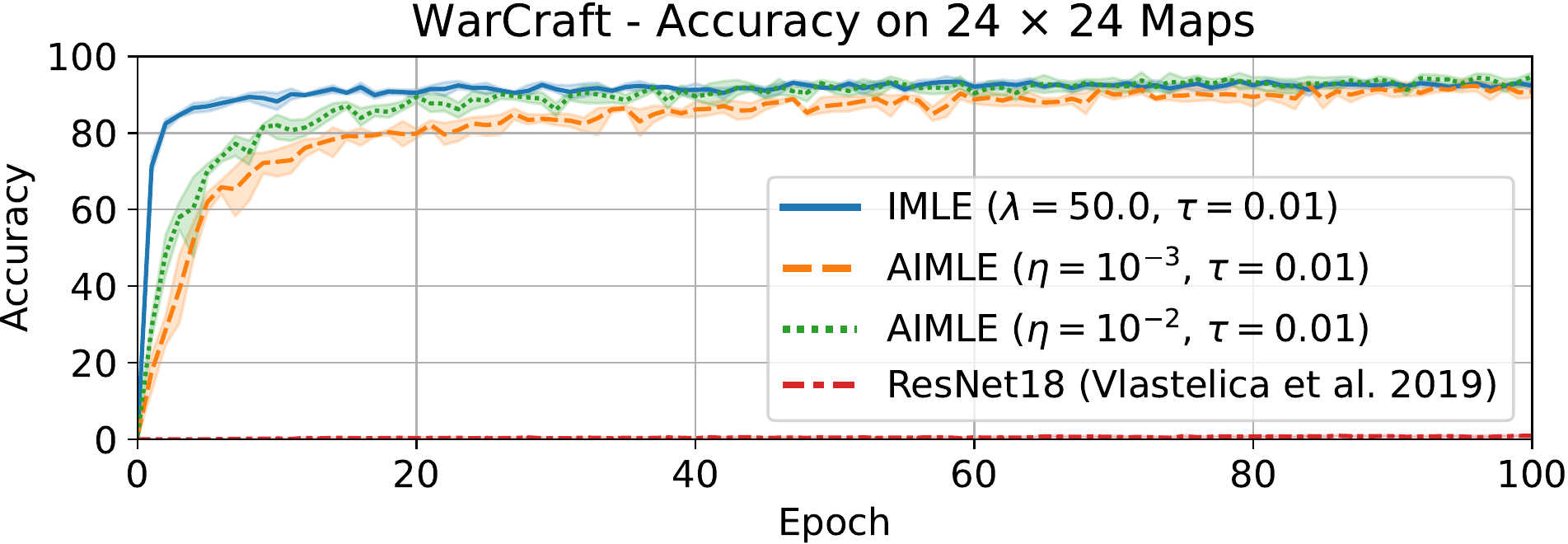}
      \includegraphics[width=1\columnwidth]{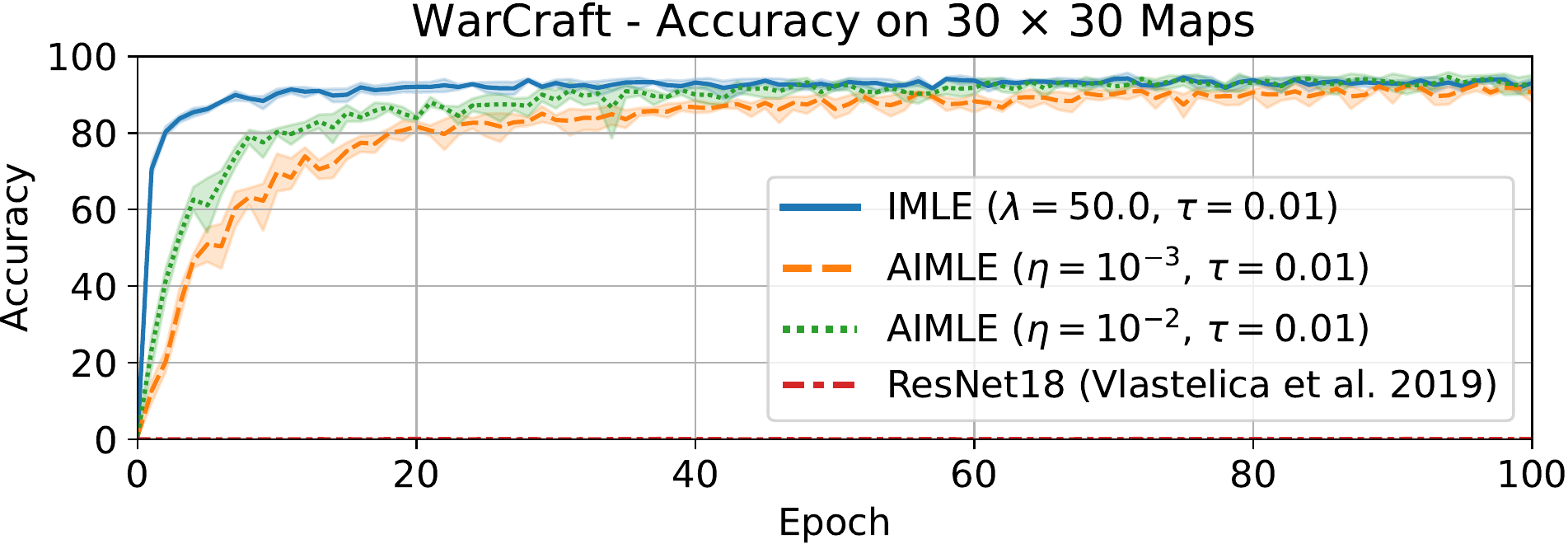}
      \caption{Training dynamics for for different models on $K \times K$ shortest path tasks on Warcraft maps, with $K \in \{ 12, 18, 24, 30 \}$.} \label{fig:w3}
\end{figure}
\paragraph{Hyper-parameters.}
We selected the hyper-parameters for SST (Spanning Tree, Hard), \imle, and \aimle, using a grid-search, and selected the hyper-parameter configuration which yields the highest ELBO on the validation set.
For SST, we searched the learning rate in $\{10^{-4}, 5 \times 10^{-4}, 10^{-3}, 5 \times 10^{-3} \}$, and the temperature in $\{ 10^{-1}, 5 \times 10^{-1}, \ldots, 10 \}$.
For \imle, we searched the learning rate in $\{10^{-4}, 5 \times 10^{-4}, 10^{-3}, 5 \times 10^{-3} \}$, $\lambda \in \{ 1, 10, 100 \}$, and the noise temperature in $\{ 0, 10^{-1}, 1, 10 \}$.
For \aimle, we searched the learning rate in $\{10^{-4}, 5 \times 10^{-4}, 10^{-3}, 5 \times 10^{-3} \}$, and the noise temperature in $\{ 0, 10^{-1}, 1, 10 \}$.
\section{Warcraft Experiments}
In these experiments, proposed by \citet{poganvcic2019differentiation}, the training datasets consist of 10,000 examples of randomly generated images of terrain maps from the Warcraft II tile set~\citep{warcraft_map_editor}.
Each example has an underlying $K \times K$ grid whose cells represent terrains with a fixed cost.
The shortest path, \ie the path with the minimum cost, between the top-left and bottom-right cells in the grid is encoded as an $K \times K$ adjacency matrix and serves as the target output.
These costs are then provided to the shortest path solver, more specifically Dijkstra's algorithm~\citep{DBLP:journals/nm/Dijkstra59}, which is used as the MAP solver to compute the shortest path between the top-left and bottom-right cells
We follow the evaluation protocol and hyper-parameters in \citet{poganvcic2019differentiation}, with the only difference that we disable the scheduled learning rate drops; the reason is that \aimle can require more time to converge since the step size $\lambda$ is initialised to $0$, and convergence can be slow for the initial part of the training process.
Our results are summarised in \cref{fig:w3}.
We can see that both \imle and its adaptive version \aimle converge to accurate models for solving the shortest path prediction tasks, producing significantly more accurate results than the ResNet18 baseline proposed by \citet{poganvcic2019differentiation}.
One limitation of \aimle is that it can require a larger number of epochs to converge: very small values of $\lambda$ at the beginning of the training procedure yield very sparse gradients, which decrease the speed of the learning process.
To this end, we evaluated two values for the update step $\eta$, namely $\eta \in \{ 10^{-3}, 10^{-2} \}$, finding that $\eta = 10^{-2}$ yields significantly higher convergence rates in comparison to the default $\eta = 10^{-3}$.

}

\end{document}